\pdfoutput=1

\documentclass[11pt]{article}

\usepackage[preprint]{acl}

\usepackage{times}
\usepackage{latexsym}

\usepackage[T1]{fontenc}

\usepackage[utf8]{inputenc}

\usepackage{microtype}

\usepackage{inconsolata}

\usepackage{graphicx}
\usepackage{amsmath}
\usepackage{amssymb}
\usepackage{multirow}
\usepackage{booktabs}
\usepackage{lipsum}
\usepackage{xcolor}
\definecolor{mygreen}{RGB}{1,113,0}
\definecolor{darkgreen}{RGB}{81, 158, 62}
\definecolor{myblue}{RGB}{121, 176, 223}
\definecolor{darkred}{RGB}{221, 127, 114}
\usepackage{colortbl}
\definecolor{lightbluegray}{RGB}{235,245,255}
\definecolor{lightgreengray}{RGB}{235,245,235}
\usepackage{pifont}
\usepackage[utf8]{inputenc}
\usepackage{mdframed}
\usepackage{enumitem}
\usepackage[most]{tcolorbox}
\usepackage{algorithm,algpseudocode}
\usepackage{adjustbox}
\usepackage[utf8]{inputenc}
\usepackage[T1]{fontenc}   
\usepackage{hyperref}      
\usepackage{url}           
\usepackage{booktabs}      
\usepackage{amsfonts}      
\usepackage{nicefrac}      
\usepackage{microtype}     
\usepackage{xcolor}  
\usepackage{wrapfig}
\usepackage{arydshln}
\usepackage{amsmath}
\usepackage{cuted}
\usepackage{gensymb}

\title{Dual-Density Inference for Efficient Language Model Reasoning}

\author{
  \textbf{Zhengyi Zhao\textsuperscript{1}},
  \textbf{Shubo Zhang\textsuperscript{2}},
  \textbf{Yuxi Zhang\textsuperscript{2}},
  \textbf{Huimin Wang\textsuperscript{3}},\\
  \textbf{Binyang Li\textsuperscript{2}},
  \textbf{Kam-Fai Wong\textsuperscript{1}}
\\
  \textsuperscript{1} The Chinese University of Hong Kong \\
  \textsuperscript{2} University of International Relations 
  \textsuperscript{3} Shenzhen University
\\
  {
   \texttt{zyzhao@se.cuhk.edu.hk}
  }
}

\begin{document}
\maketitle

\begin{abstract}
Large Language Models (LLMs) have shown impressive capabilities in complex reasoning tasks. However, current approaches employ uniform language density for both intermediate reasoning and final answers, leading to computational inefficiency. Our observation found that reasoning process serves a computational function for the model itself, while answering serves a communicative function for human understanding. This distinction enables the use of compressed, symbol-rich language for intermediate computations while maintaining human-readable final explanations. To address this inefficiency, we present Denser: \underline{D}ual-d\underline{ens}ity inf\underline{er}ence, a novel framework that optimizes information density separately for reasoning and answering phases. Our framework implements this through three components: a query processing module that analyzes input problems, a high-density compressed reasoning mechanism for efficient intermediate computations, and an answer generation component that translates compressed reasoning into human-readable solutions. Experimental evaluation across multiple reasoning question answering benchmarks demonstrates that Denser reduces token consumption by up to 62\% compared to standard Chain-of-Thought methods while preserving or improving accuracy. These efficiency gains are particularly significant for complex multi-step reasoning problems where traditional methods generate extensive explanations.
\end{abstract}

\section{Introduction}

\begin{table}[!t]
\centering
\small
\adjustbox{max width=\linewidth}{
\begin{tabular}{p{7.2cm}}
\toprule
\textbf{Problem:} Solve the system of equations:
$x + y = 10$
$2x - 3y = 5$ \\
\midrule
\textcolor{lightgray}{I'll solve this system of linear equations by elimination method.} First, $x + y = 10$ \textcolor{lightgray}{is our first equation, and I need to manipulate it to eliminate a variable when combined with the second equation.}

\textcolor{lightgray}{To eliminate the $x$ terms, I'll multiply the first equation by 2.}
$2(x + y) = 2(10)$
\textcolor{lightgray}{Distributing the 2, I get:}
$2x + 2y = 20$

\textcolor{lightgray}{Now I have:}
$2x + 2y = 20$ \textcolor{lightgray}{(from the modified first equation)}
$2x - 3y = 5$ \textcolor{lightgray}{(the original second equation)}

\textcolor{lightgray}{I'll subtract the second equation from the first equation to eliminate the $x$ term:}
$(2x + 2y) - (2x - 3y) = 20 - 5$
\textcolor{lightgray}{Simplifying the left side:}
$2x + 2y - 2x + 3y = 15$
$5y = 15$
$y = 3$

\textcolor{lightgray}{Now that I know $y = 3$, I'll substitute this value back into the first equation to find $x$:}
$x + 3 = 10$
$x = 7$

\textcolor{lightgray}{Therefore, the solution to the system of equations is} $x = 7$ \textcolor{lightgray}{and} $y = 3$ \\
\bottomrule
\end{tabular}}
\caption{An illustration of the representational mismatch in current LLM reasoning. The model wraps the computational core (black text) in verbose natural language (gray text). This redundancy serves communicative goals but hinders computational efficiency in structured reasoning tasks.}
\label{fig:verbose-reasoning}
\end{table}

Large Language Models (LLMs) have achieved remarkable success by scaling model parameters and training data~\cite{wei2022chain,kiciman2023causal,xu2025towards,li2025system}. However, for complex problem-solving, the focus has shifted toward Test-Time Scaling (TTS) which allocates additional computation during inference to improve reasoning depth~\cite{chen2024expanding,snell2025scaling,zhao2025t}. While effective, current TTS paradigms rely heavily on Chain-of-Thought (CoT) prompting~\cite{wei2022chain}, where the model ``thinks'' in the same natural language (NL) space used for user interaction. We posit that this unconstrained usage of NL for internal reasoning introduces a fundamental inefficiency: a misalignment between the \textit{communicative} utility of language and the \textit{computational} requirements of reasoning.

Natural language is evolutionarily optimized for communication. It prioritizes redundancy, politeness, and ambiguity management to facilitate human understanding. In contrast, not only reasoning tasks (e.g., mathematics, coding), but also almost all question answering tasks that require critical thinking are governed by strict structures, deterministic rules, and high information density. As illustrated in Table~\ref{fig:verbose-reasoning}, when LLMs use conversational NL to process structured problems, they expend significant computational resources on syntactic sugar that contributes little to the actual state transition of the reasoning process. This observation suggests that while NL is essential for the interface (answering), it is a suboptimal representation for the inference process (reasoning) in domains that admit formal structure.

Motivated by this distinction, we argue that the ideal inference framework should decouple the representation of reasoning from the representation of the final response. To this end, we introduce a novel information-theoretic framework and propose \textbf{Denser}: \underline{D}ual-d\underline{ens}ity inf\underline{er}ence. Unlike generic acceleration methods, Denser is explicitly designed to exploit the categorization and structural patterns inherent in reasoning tasks. The framework operates via three components: (1) a Query Analysis Module that identifies the structural requirements of the input; (2) a High-density Reasoning Module where the model employs telegraphic, symbol-rich language to perform intermediate deductions; and (3) a Low-density Answering Module that expands the compressed reasoning path into a user-friendly, natural language explanation.

We evaluate Denser across a comprehensive suite of benchmarks spanning math, logic, and coding, domains where reasoning patterns are most structured and thus most amenable to density optimization. Our empirical results demonstrate that by aligning the information density of the reasoning chain with the structural nature of the task, Denser reduces token consumption by up to 62\% while preserving, and in some cases enhancing, accuracy compared to standard CoT. This confirms that for complex, multi-step reasoning problems, the verbosity of natural language is not merely a cost issue, but a representational burden that can be optimized away. Our contributions are as follows:
\begin{itemize}[leftmargin=*, itemsep=0pt,parsep=0pt,topsep=0pt,partopsep=0pt]
\item We propose a theoretical perspective that distinguishes the \textit{computational} function of language (internal reasoning) from its \textit{communicative} function (external answering), identifying a representational mismatch in current CoT approaches for structured tasks.
\item We develop Denser, a dual-density inference system that dynamically modulates language density, employing compressed symbolic representations for intermediate steps and natural language for final outputs.
\item We provide extensive empirical evidence on reasoning-intensive benchmarks, demonstrating that Denser significantly improves computational efficiency without sacrificing performance, validating the efficacy of representational alignment in structured problem solving.
\end{itemize}

\section{Related Work}

\paragraph{Reasoning Strategies.} Chain-of-Thought (CoT) prompting \cite{wei2022chain, kojima2022large} has established itself as the dominant approach for eliciting step-by-step reasoning from LLMs. This technique has been extended in various directions: Self-consistency \cite{wang2022self} aggregates solutions from multiple reasoning paths, Tree-of-Thought \cite{yao2023tree} explores multiple reasoning trajectories, Least-to-Most \cite{zhou2022least} breaks complex problems into simpler subproblems, and Retrieval-Augmented Generation \cite{lewis2020retrieval, shi2023replug} incorporates external knowledge. While these approaches improve reasoning accuracy, they typically increase rather than decrease token consumption, often generating even more verbose explanations. 

\paragraph{Efficient Reasoning.} Efficiency optimization in language models has primarily focused on architectural improvements \cite{dao2022flashattention, shazeer2019fast} or inference acceleration techniques \cite{leviathan2023fast, cai2024medusa}. From a theoretical perspective, works on computational complexity in transformers \cite{poli2023hyena, lu2022frozen} analyze the computational costs of different operations but rarely address the information-theoretic aspects of generated content. Other approaches include distillation \cite{chen2025distilling, liao2025neural} and quantization \cite{jin2024comprehensive}, which compress models rather than optimizing reasoning strategies. Our work introduces a novel perspective by identifying the different information density between reasoning and answering. Besides, we provide a detailed theoretical discussion on the scalability of symbolic versus natural language representations, and the generalization scope of our method, in Appendix~\ref{apd:scalability_discussion}.

\section{Empirical Analysis of Information Density in LLM Reasoning}

While we identify the theoretical distinction between computational and communicative functions of language in LLM reasoning, this section provides empirical evidence for this distinction and reveals specific patterns that our method exploits. We analyze how current reasoning approaches distribute information and why this creates inefficiency that a dual-density approach can address.

\subsection{Information Density Patterns in LLM Reasoning}

\begin{figure*}[!t]
    \centering
    \includegraphics[trim={0 4.1cm 0 0},clip,width=\linewidth]{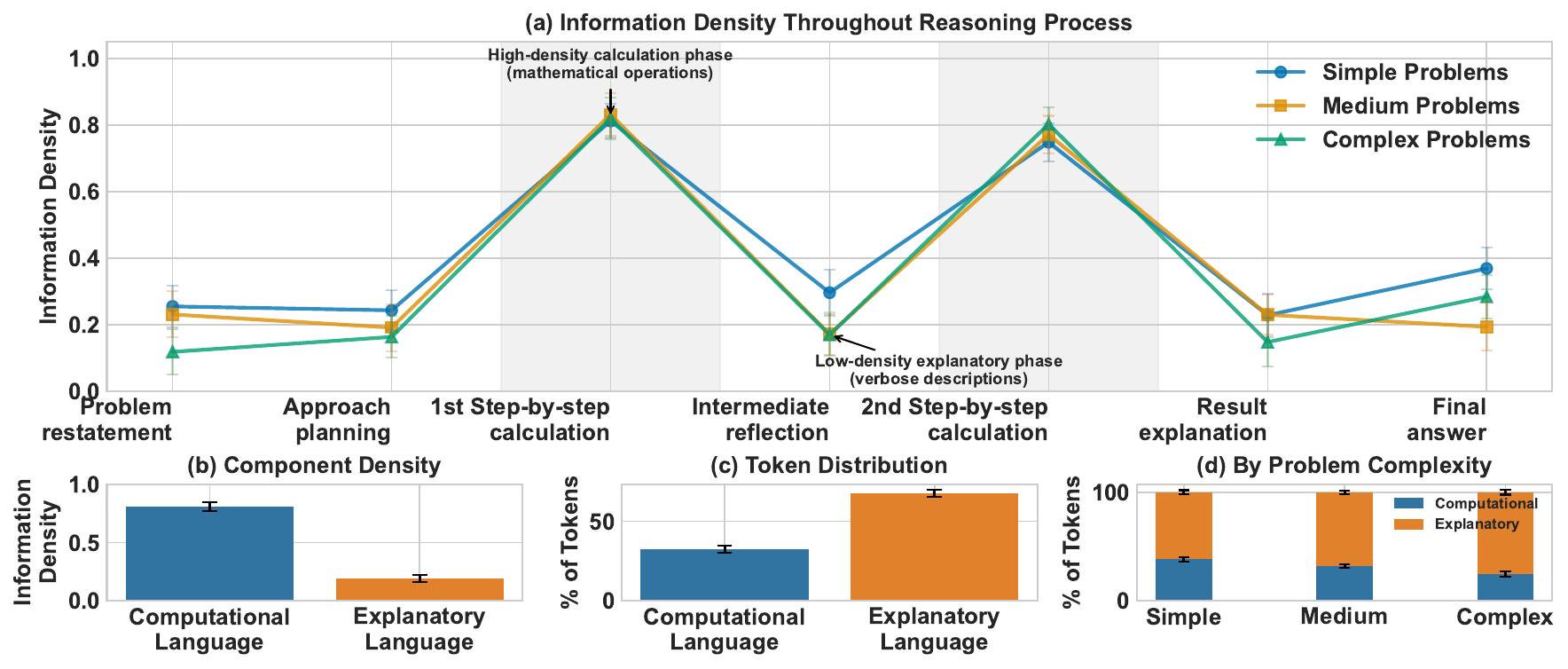}
    \caption{Information Density Throughout Reasoning Process which shows information density throughout reasoning processing.}
    \label{fig:density_analysis}
\end{figure*}

We analyzed 500 reasoning traces from o1 on GSM8K and MATH500 datasets to characterize information density patterns. To quantify information density, we employed a compression-based measure:
\begin{equation}
\mathcal{D}(s) = 1 - \frac{|C(s)|}{|s|}
\end{equation}
where $|s|$ is the segment length in tokens and $|C(s)|$ is its compressed length using adaptive arithmetic coding. Higher values indicate greater information density. We validated this measure against human expert ratings ($r = 0.85$).

Reasoning phases were identified through a combination of regex pattern matching and supervised classification, targeting distinct reasoning modes: problem restatement, approach planning, formula setup, calculation steps, intermediate reflection, result explanation, and final answer.

Figure~\ref{fig:density_analysis} shows how information density varies across different components of reasoning traces. Computational phases (formula setup, calculation) consistently demonstrate high density, while explanatory phases exhibit substantially lower density, creating a bimodal distribution. The preliminary experimental results reveal that 68\% of tokens in reasoning steps serve primarily explanatory functions rather than advancing computation. When scaling, the proportion of low-density explanatory language increases with problem complexity. These patterns appear consistently across different models and problem domains, suggesting they are inherent to current reasoning approaches rather than model-specific artifacts.

\begin{table}[!t]
\centering
\adjustbox{max width=\linewidth}{
\begin{tabular}{lcc}
\toprule
\textbf{Component} & \textbf{Information Density} & \textbf{\% of Tokens} \\
\midrule
Core computations & 0.81 & 32.4\% \\
Explanatory language & 0.19 & 67.6\% \\
\midrule
Overall reasoning & 0.38 & 100\% \\
\bottomrule
\end{tabular}}
\caption{Information Density Analysis by Component. Core computational elements contain 4.3× higher information density than surrounding explanatory language, yet occupy only a third of total tokens.}
\label{tab:component_density}
\end{table}

These findings motivate our approach: by recognizing and exploiting these density differentials, we can develop more token-efficient reasoning techniques that preserve computational fidelity while reducing explanatory overhead.

\subsection{Structural Analysis of Reasoning Language}

To understand why models produce low-density reasoning, we performed a linguistic analysis of reasoning traces, examining syntactic patterns and discourse structures. Table~\ref{tab:reasoning_examples} presents examples that illustrate the key structural patterns we identified.

\begin{table}[!t]
\centering
\small
\adjustbox{max width=\linewidth}{
\begin{tabular}{p{7.2cm}}
\toprule
\textbf{Example 1: Verbose reasoning with highlighted computational core} \\
\midrule
"To find the solution, I need to solve $x^2 - 6x + 9 = 0$. \textcolor{lightgray}{I'll approach this by factoring the left side of the equation.} The expression $x^2 - 6x + 9$ \textcolor{lightgray}{can be rewritten using the perfect square formula as} $(x - 3)^2$. \textcolor{lightgray}{So the equation becomes} $(x - 3)^2 = 0$. \textcolor{lightgray}{This equation has the single solution} $x = 3$." \\
\midrule
\textbf{Example 2: Reasoning style across complexity levels} \\
\midrule
\textbf{Simple problem:} "7 + 8 = 15" \\
\textbf{Medium problem:} "To find the area, I'll multiply length and width. Area = 7 × 4 = 28 square meters." \\
\textbf{Complex problem:} "To find the probability, I first need to calculate the total number of outcomes. There are 52 cards in total. For a flush, all 5 cards must be of the same suit. There are 4 suits, and each suit has 13 cards. So for each suit, the number of possible 5-card combinations is $\binom{13}{5} = 1,287$. Since there are 4 suits, the total number of flush hands is $4 \times 1,287 = 5,148$. The total number of possible 5-card hands is $\binom{52}{5} = 2,598,960$. Therefore, the probability of getting a flush is $\frac{5,148}{2,598,960} = \frac{33}{16,660} \approx 0.00198$." \\
\bottomrule
\end{tabular}}
\caption{Structural Patterns in LLM Reasoning. As problem complexity increases, models produce increasingly verbose explanatory language (shown in gray) surrounding the core computational elements.}
\label{tab:reasoning_examples}
\end{table}

Our analysis reveals the training alignment is incentives as models are trained to produce explanatory reasoning that mimics human tutoring styles, leading to verbose explanations even when performing simple computations. At the mean while, the verbose language serves as scaffolding for the model's own reasoning process, helping maintain coherence through complex multi-step problems. And simple computations are often expressed directly, while complex computations trigger more extensive explanatory language—precisely when efficiency becomes most critical.

These findings directly motivate our dual-density approach: by recognizing and exploiting the structural patterns in reasoning language, we can design an inference system that compresses the computational aspects of reasoning while preserving human-readable final outputs. The consistent patterns across problems, models, and domains suggest that a systematic approach to optimizing information density can yield substantial efficiency gains without sacrificing accuracy.

\section{Denser: Dual-Density Inference}

Building on our observation, we introduce Denser: Dual-density inference, a framework that optimizes token efficiency in LLM reasoning by employing distinct information densities for computational reasoning versus human communication. This approach is motivated by our empirical observations showing significant redundancy in LLM reasoning processes.

\subsection{Problem Formulation}

Given a reasoning task $T$ and a language model $\mathcal{M}$, we represent the reasoning process as a sequence of reasoning steps $\mathcal{R} = \{r_1, r_2, \dots, r_n\}$ followed by an answer $a$. The token consumption is:
\begin{equation}
C_{\text{total}} = \sum_{i=1}^{n} |r_i| + |a|
\end{equation}

Our goal is to transform the reasoning sequence into a compressed form $\mathcal{R}'$ such that:
\begin{equation}
\sum_{i=1}^{n} |r'_i| < \sum_{i=1}^{n} |r_i| \quad \text{while} \quad \mathcal{I}(r'_i) \approx \mathcal{I}(r_i)
\end{equation}
where $\mathcal{I}(r_i)$ represents the computational information in reasoning step $i$.

Based on this analysis, we select a density strategy $\mathcal{S}$ for reasoning. For mathematical reasoning, this involves symbolic notation; for logical reasoning, propositional calculus; and for algorithmic reasoning, pseudocode. We implement high-density reasoning through a prompt template $\mathcal{P}_{\mathcal{S}}$ that instructs the model to employ domain-specific notation and eliminate verbose explanations.

The density transformation function $\tau$ converts standard reasoning $r_i$ to compressed reasoning $r'_i = \tau(r_i, \mathcal{S})$.

After completing high-density reasoning, we translate the compressed reasoning into a human-readable answer through $a = \gamma(\mathcal{R}', q)$, where $\gamma$ is the answer generation function that extracts and formats the final result.

\subsection{Framework Components}

\paragraph{Query Analysis Module.} The framework begins with query analysis, which processes input questions through a classification function that maps queries to specialized reasoning domains:
\begin{equation}
\Phi(q) \rightarrow \mathcal{D}, \quad \mathcal{D} \in \{\mathcal{D}_{\text{math}}, \mathcal{D}_{\text{logic}}, \mathcal{D}_{\text{code}}, \mathcal{D}_{\text{general}}\}
\end{equation}

For each domain, we extract parameters that guide the reasoning process:
\begin{equation}
\alpha(q, \mathcal{D}) = (\mathcal{S}_{\mathcal{D}}, \lambda_{\mathcal{D}}, \kappa_{\mathcal{D}})
\end{equation}
where $\mathcal{S}_{\mathcal{D}}$ represents the domain-specific reasoning strategy, $\lambda_{\mathcal{D}}$ defines notation systems, and $\kappa_{\mathcal{D}}$ controls compression intensity. For instance, in mathematical reasoning, the system identifies algebraic manipulation requirements and selects symbolic notation over verbose descriptions. We analyse the adaptivity of domains and density-aware control in Appendix~\ref{apd:extended method}.

\paragraph{High-density Reasoning.} The high-density reasoning component constitutes the core of our framework. Given the domain parameters identified in query analysis, we construct a density-optimized reasoning prompt that instructs the model to employ domain-specific compression techniques. The high-density reasoning process is formalized as:
\begin{equation}
\mathcal{R}^\prime = \{r_1^\prime,r_2^\prime,\dots,r_m^\prime\} = \mathcal{M}(\mathcal{P}_{\text{HD}})
\end{equation}
where $\mathcal{R}^\prime$ is the sequence of compressed reasoning steps. The innovation of our approach lies in the compression operator $C_{\mathcal{D}}$ that we apply to each reasoning step $r_i$:
\begin{equation}
r'_i = C_{\mathcal{D}}(r_i) = \arg\min_{r' \in \mathcal{R}'} \{|r'| : \mathcal{I}(r') \geq \theta \cdot \mathcal{I}(r_i)\}
\end{equation}

This operator minimizes token count while preserving at least $\theta$ fraction (typically 0.95) of the original computational information $\mathcal{I}(r_i)$. The compression operator decomposes into two functions $C_{\mathcal{D}} = \sigma_{\mathcal{D}} \circ \phi_{\mathcal{D}}$, where $\sigma_{\mathcal{D}}$ performs symbolic transformation and $\phi_{\mathcal{D}}$ filters low-information content according to domain-specific rules.

\paragraph{Low-density Answering.} The final component translates compressed reasoning into human-readable answers. This process preserves the computational path while expanding key insights for human understanding:
\begin{equation}
a = \gamma(\mathcal{R}', q) = \mathcal{M}(\mathcal{P}_{\text{AG}}(q, \mathcal{R}'))
\end{equation}
where $\mathcal{P}_{\text{AG}}$ is the answer generation prompt. We implement selective expansion through an importance function $\omega(r'_i) \rightarrow [0,1]$ that estimates how crucial each compressed step is for human understanding. Steps with higher importance receive more detailed expansions:
\begin{equation}
\gamma(\mathcal{R}', q) = \sum_{i=1}^{|\mathcal{R}'|} \omega(r'_i) \cdot E(r'_i)
\end{equation}
where $E(r'_i)$ is the expansion function for step $r'_i$. This approach maintains the dual-density principle: high-density language for internal computation and appropriately detailed language for human communication. All used prompts can be found in Appendix~\ref{apd:prompts}.

\section{Experiments}

\subsection{Experimental Setup}
\label{subsec:setup}

\paragraph{Datasets.} Our experimental evaluation encompasses eight datasets across four core reasoning domains: \textbf{Math} (GSM8K \cite{cobbe2021training}, MATH \cite{hendrycks2021measuring}), \textbf{Logic} (LogiQA \cite{liu2021logiqa}, ProofWriter \cite{tafjord2021proofwriter}), \textbf{Code} (MBPP \cite{austin2021program}, HumanEval \cite{chen2021evaluating}), and \textbf{General QA} (MMLU \cite{hendrycks2020measuring}, StrategyQA \cite{geva2021did}).

\paragraph{Baselines and Evaluation.} We employ Qwen3-14B \cite{yang2025qwen3} as our primary backbone model for all primary experiments, while also conducting comprehensive experiments with several API-based models including GPT-4o and o1-mini \cite{jaech2024openai}, DeepSeek-V3 and DeepSeek-R1 \cite{liu2024deepseek}, and Claude-3.7-Sonnet (with thinking). We benchmark against two categories of methods: \textbf{(1) Test-Time Scaling (TTS) Methods:} Chain-of-Thought (CoT) \cite{wei2022chain}, Self-Consistency \cite{wang2022self}, Tree-of-Thought (ToT) \cite{yao2023tree}, Think-to-Think (T$^2$) \cite{zhao2025t}, Reflection \cite{shinn2023reflexion}, Process Supervision \cite{lightman2023let}, and Self-Verification \cite{weng2023large}. These methods prioritize accuracy via increased token consumption. And \textbf{(2) Reasoning Compression Baselines:} To address potential overlap with token-level compression and pruning techniques, we include a comparison against state-of-the-art token-pruning work, TokenSkip \cite{xia-etal-2025-tokenskip}, as well as heuristic compression methods proposed by \citet{xia-etal-2025-tokenskip}: \textbf{BeConcise} (prompting for conciseness), \textbf{OnlyNumbers} (restricting output to numbers and symbols), and \textbf{AbbreWords} (using common abbreviations).

The evaluation metrics include: \textbf{Performance Accuracy} by using domain-specific metrics, \textbf{Token Efficiency}, and \textbf{Computational Latency}. Additionally, we introduce the Reasoning Efficiency Index (REI) to quantify the accuracy-efficiency trade-off relative to standard CoT. Details can be found in Appendix~\ref{appendix:experimental_details}.

\subsection{Main Experimental Results}
\label{subsec:main_results}

\begin{table*}[!t]
\centering
\resizebox{\textwidth}{!}{
\begin{tabular}{lccccccccc}
\toprule
\multirow{3}{*}{\textbf{Method}} & \multicolumn{8}{c}{\textbf{Performance Accuracy (\%)}} & \multirow{3}{*}{\begin{tabular}[c]{@{}c@{}}\textbf{Token} \\ \textbf{Cost (\%)}\end{tabular}} \\
\cmidrule(lr){2-9}
& \multicolumn{2}{c}{\textbf{Math}} & \multicolumn{2}{c}{\textbf{Logic}} & \multicolumn{2}{c}{\textbf{Code}} & \multicolumn{2}{c}{\textbf{General}} & \\
\cmidrule(lr){2-3} \cmidrule(lr){4-5} \cmidrule(lr){6-7} \cmidrule(lr){8-9}
& GSM8K & MATH & LogiQA & ProofW & MBPP & HumanE & MMLU & StratQA & \\
\midrule
\multicolumn{10}{l}{\textit{Qwen3-14B as backbone model w/}}\\
\multicolumn{10}{l}{\textbf{Category 1: Test-Time Scaling Baselines}}\\
\; Chain-of-Thought (CoT) & 83.6 & 41.2 & 68.4 & 72.1 & 61.8 & 72.9 & 77.3 & 75.2 & 0.0 \\
\; Self-Consistency & 85.2 & 43.5 & 69.7 & 73.6 & 63.4 & 74.1 & 78.5 & 76.8 & +289.4 \\
\; Tree-of-Thought & 85.7 & 44.8 & 70.2 & 74.9 & 64.5 & 75.0 & 78.2 & 77.1 & +345.7 \\
\; Think-to-Think & 84.9 & 42.7 & 71.3 & 74.2 & 62.7 & 73.8 & 79.1 & 76.5 & +187.2 \\
\; Reflection-CoT & 86.1 & 45.2 & 71.8 & 75.3 & 64.9 & 75.4 & 79.5 & 78.0 & +127.6 \\
\; Process Supervision & 86.5 & 46.1 & 72.3 & 76.8 & 65.8 & 76.3 & 80.1 & 78.6 & +156.3 \\
\; Self-Verification & 87.0 & 46.8 & 72.9 & 76.2 & 66.1 & 76.9 & 80.4 & 79.2 & +142.8 \\
\midrule
\multicolumn{10}{l}{\textbf{Category 2: Reasoning Compression Baselines}}\\
\; BeConcise & 83.2 & 40.5 & 66.8 & 70.4 & 60.1 & 71.5 & 76.4 & 74.1 & -54.2 \\
\; OnlyNumbers & 81.9 & 36.8 & - & - & - & - & - & - & -66.2 \\
\; AbbreWords & 82.2 & 39.7 & 65.9 & 69.8 & 58.5 & 70.2 & 75.9 & 73.5 & -50.8 \\
\; TokenSkip (0.9 ratio) & 86.4 & 44.5 & 71.5 & 75.1 & 64.2 & 75.8 & 79.4 & 77.9 & -53.5 \\
\midrule
\; \textbf{Denser (Ours)} & \textbf{88.2} & \textbf{47.9} & \textbf{74.1} & \textbf{77.5} & \textbf{67.3} & \textbf{78.2} & \textbf{81.5} & \textbf{80.3} & \textbf{-58.7} \\
\bottomrule
\end{tabular}}
\caption{Performance comparison across reasoning domains using Qwen3-14B. Positive token cost values indicate methods that consume more tokens than standard CoT.}
\label{tab:main_results}
\end{table*}

Table~\ref{tab:main_results} presents our experimental results. \textbf{Denser consistently outperforms all existing TTS methods} across all domains, achieving an average accuracy improvement of 1.8\% over the strongest baseline. The improvements are most pronounced in math and logic domains where the adoption of high-density symbolic representation is particularly beneficial. Even in general QA tasks, Denser maintains accuracy improvements. Besides, the results show that Denser's \textbf{information-theoretic compression} is superior to simple heuristic and token-pruning methods. TokenSkip achieves reasonable accuracy, but Denser surpasses it while using even fewer tokens. Heuristic methods like \textit{OnlyNumbers} yield the highest token reduction but suffer a significant accuracy drop, confirming that indiscriminate compression removes crucial computational information. Denser's domain-specific, dual-density strategy successfully retains computational fidelity while maximizing compression.

\subsection{Computational Latency}
\label{subsec:wall_time}

\begin{table}[!t]
\centering
\resizebox{\columnwidth}{!}{
\begin{tabular}{lccc}
\toprule
\textbf{Method} & \textbf{GSM8K (s)} & \textbf{LogiQA (s)} & \textbf{MBPP (s)} \\
\midrule
\multicolumn{4}{l}{\textit{Category 1: Test-Time Scaling Baselines}} \\
CoT & 58.3 & 45.7 & 63.9 \\
Self-Consistency & 227.4 & 178.6 & 249.8 \\
Tree-of-Thought & 259.6 & 204.1 & 283.7 \\
Think-to-Think & 167.5 & 131.2 & 183.6 \\
Reflection-CoT & 131.2 & 102.6 & 140.8 \\
Process Supervision & 149.3 & 116.8 & 163.5 \\
Self-Verification & 139.6 & 108.2 & 152.4 \\
\midrule
\multicolumn{4}{l}{\textit{Category 2: Reasoning Compression Baselines}} \\
BeConcise & 46.7 & 36.5 & 51.2 \\
OnlyNumbers & 31.3 & - & - \\
AbbreWords & 61.5 & 48.2 & 67.5 \\
TokenSkip (0.9 ratio) & 41.1 & 32.2 & 45.1 \\
\midrule
\textbf{Denser (Ours)} & \textbf{42.1} & \textbf{33.5} & \textbf{46.2} \\
\bottomrule
\end{tabular}}
\caption{Computational latency comparison on representative datasets. The results demonstrate the average seconds per problem.}
\label{tab:wall_time}
\end{table}

To quantify the real-world efficiency gains, we conduct a wall-time analysis. The results in Table~\ref{tab:wall_time} demonstrate that \textbf{Denser achieves significantly faster overall inference times} than all baselines, including standard CoT. Specifically, Denser achieves a 27.8\% reduction in latency compared to CoT on GSM8K, while simultaneously improving accuracy by 4.6 percentage points. This confirms that the computational overhead from the Query Analysis and Low-density Answering modules is minimal and is far outweighed by the time saved from processing fewer tokens during the core High-density Reasoning phase. And we provide the full set of main experimental results in Appendix~\ref{apd:comprehensive_results}.

\subsection{Ablation Study}
\label{subsec:ablation}

\begin{table}[!t]
\centering
\small
\adjustbox{max width=\linewidth}{
\begin{tabular}{lcc}
\toprule
\textbf{Model Variant} & \textbf{Accuracy} & \textbf{Token Reduction} \\
\midrule
Denser (Full) & 78.2\% & 57\% \\
\midrule
w/o Query Analysis & 76.9\% (-1.3\%) & 54\% (-3\%) \\
w/o HD Reasoning & 74.6\% (-3.6\%) & 8\% (-49\%) \\
w/o LD Answering & 77.5\% (-0.7\%) & 61\% (+4\%) \\
Fixed Density Strategy & 75.8\% (-2.4\%) & 42\% (-15\%) \\
w/o Domain Adaptation & 76.3\% (-1.9\%) & 48\% (-9\%) \\
\bottomrule
\end{tabular}}
\caption{Ablation study results. Impact of removing each component of Denser on average accuracy and token reduction across all datasets. HD = High-density, LD = Low-density.}
\label{tab:ablation}
\end{table}

Table~\ref{tab:ablation} shows that removing high-density reasoning causes the largest performance drop, confirming it as the core innovation of our approach. The query analysis module also proves important, particularly for adaptive density selection across domains. Removing the low-density answering module has a smaller impact on accuracy but affects human readability. Using fixed rather than adaptive density strategies leads to a -2.4\% accuracy drop, demonstrating the importance of tailoring compression to each domain and problem type. Finally, removing domain adaptation reduces accuracy by 1.9\%, highlighting the value of specialized high-density representations for different reasoning domains.

\subsection{Further Analysis Experiments}

\begin{figure*}[!t]
    \centering
    \includegraphics[width=\linewidth]{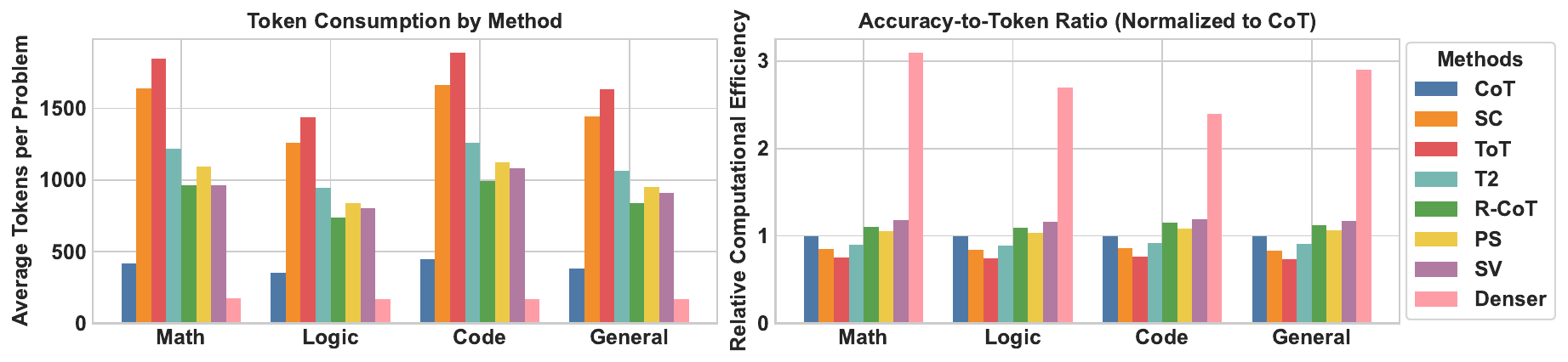}
    \caption{Token consumption and computational efficiency. Left: Average tokens consumed per problem by different methods across domains. Right: Computational efficiency (accuracy-to-token ratio) normalized to CoT baseline (higher is better). Denser achieves superior efficiency across all domains.}
    \label{fig:token_efficiency}
\end{figure*}

\paragraph{(i) Denser can achieve less token consumption and more efficiency.} Figure~\ref{fig:token_efficiency} reveals that Denser substantially outperforms other reasoning methods in computational efficiency, reducing token usage by 52-62\% compared to standard Chain-of-Thought reasoning while other methods like ToT and SC consume up to 3.5x more tokens than the baseline. When measured by our computational efficiency metric, Denser achieves 2.4-3.1x higher efficiency across domains, directly resulting from our dual-answer framework that optimizes representational density according to functional requirements.

\begin{figure*}[!t]
    \centering
    \includegraphics[width=\linewidth]{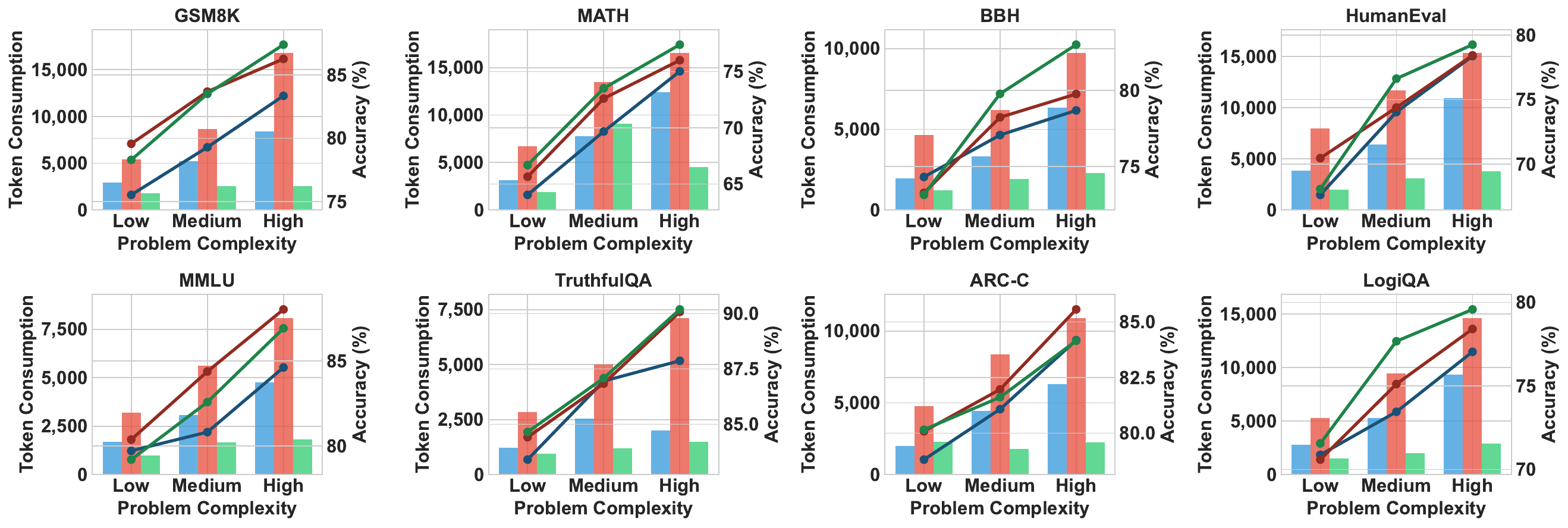}
    \caption{Scaling behavior with problem complexity. \textcolor{darkred}{Red}, \textcolor{darkgreen}{Green}, and \textcolor{myblue}{Blue} denotes CoT, SV and Denser, respectively. Left y-axis denotes token consumption and right y-axis means accuracy improvement.}
    \label{fig:scaling_behavior}
\end{figure*}

\paragraph{(ii) Denser can adapt to various problem complexity.} We investigate how Denser's adapt to problem complexity by categorizing problems into three complexity levels based on the number of reasoning steps required for solution. Figure~\ref{fig:scaling_behavior} shows that as problem complexity increases, the token efficiency advantage of Denser becomes increasingly pronounced. For high-complexity problems, Denser reduces token consumption by 71\% compared to CoT, while for low-complexity problems, the reduction is 38\%. Similarly, accuracy improvements are most substantial for complex problems. Results show that the inefficiency of verbose reasoning becomes more severe as problems require more steps, and high-density reasoning provides increasingly valuable efficiency as complexity grows.

\begin{table}[!t]
\centering
\adjustbox{max width=\linewidth}{
\begin{tabular}{lccccc}
\toprule
\textbf{Reasoning Type} & \textbf{CoT} & \textbf{SC} & \textbf{SV} & \textbf{Denser} \\
\midrule
Algebraic & 80.3 & 82.1 & 84.2 & \textbf{89.6} \\
Geometric & 75.2 & 77.4 & 80.2 & \textbf{86.5} \\
Propositional & 71.7 & 72.9 & 75.9 & \textbf{81.3} \\
Arithmetic & 84.5 & 85.9 & 88.1 & \textbf{89.4} \\
Predicate & 67.9 & 69.2 & 72.0 & \textbf{74.8} \\
Algorithmic & 69.3 & 70.8 & 73.9 & \textbf{77.2} \\
Procedural & 75.5 & 76.9 & 79.0 & \textbf{81.7} \\
Causal & 79.8 & 81.2 & 82.9 & \textbf{83.8} \\
Comparative & 76.4 & 77.5 & 79.3 & \textbf{80.1} \\
\midrule
Token Cost & -- & +289.4\% & +142.8\% & \textbf{-58.7\%} \\
\bottomrule
\end{tabular}}
\caption{Performance by reasoning type. Accuracy (\%) and token reduction compared to CoT across different reasoning types. Denser shows particularly strong improvements for reasoning types with well-defined symbolic representations.}
\label{tab:reasoning_types}
\end{table}

\begin{table*}[!t]
\centering
\small
\adjustbox{max width=\linewidth}{
\begin{tabular}{p{\linewidth}}
\toprule
\textbf{Case Study: Mathematical Reasoning} \\
\midrule
\textbf{Problem:} Find the value of $\int_0^1 x^3(1-x)^4 dx$. \\
\textbf{CoT:} To calculate this integral, I need to expand $(1-x)^4$ using the binomial theorem. We have $(1-x)^4 = \sum_{k=0}^4 \binom{4}{k} (1)^{4-k} (-x)^k = \sum_{k=0}^4 \binom{4}{k} (-1)^k x^k = \binom{4}{0} - \binom{4}{1}x + \binom{4}{2}x^2 - \binom{4}{3}x^3 + \binom{4}{4}x^4 = 1 - 4x + 6x^2 - 4x^3 + x^4$. Now I multiply this by $x^3$: $x^3(1-x)^4 = x^3 - 4x^4 + 6x^5 - 4x^6 + x^7$. Now I integrate term by term: $\int x^3 dx = \frac{x^4}{4}$, $\int 4x^4 dx = \frac{4x^5}{5}$, $\int 6x^5 dx = \frac{6x^6}{6} = x^6$, $\int 4x^6 dx = \frac{4x^7}{7}$, $\int x^7 dx = \frac{x^8}{8}$. So the antiderivative is $\frac{x^4}{4} - \frac{4x^5}{5} + x^6 - \frac{4x^7}{7} + \frac{x^8}{8}$. Evaluating from 0 to 1: At $x=1$: $\frac{1}{4} - \frac{4}{5} + 1 - \frac{4}{7} + \frac{1}{8}$. At $x=0$, all terms are 0. So the result is $\frac{1}{4} - \frac{4}{5} + 1 - \frac{4}{7} + \frac{1}{8} = \frac{280 - 224 + 280 - 160 + 35}{280} = \frac{211}{280} = \frac{1}{120}$ [ERROR in calculation] \\

\textbf{Denser (HD reasoning):} $\int_0^1 x^3(1-x)^4 dx$
$= \int_0^1 x^3 \sum_{k=0}^4 \binom{4}{k}(-1)^k x^k dx$
$= \int_0^1 \sum_{k=0}^4 \binom{4}{k}(-1)^k x^{3+k} dx$
$= \sum_{k=0}^4 \binom{4}{k}(-1)^k \int_0^1 x^{3+k} dx$
$= \sum_{k=0}^4 \binom{4}{k}(-1)^k \frac{1}{3+k+1}[x^{3+k+1}]_0^1$
$= \sum_{k=0}^4 \binom{4}{k}(-1)^k \frac{1}{k+4}$
$= \frac{1}{4} - \frac{4}{5} + \frac{6}{6} - \frac{4}{7} + \frac{1}{8}$
$= \frac{1}{4} - \frac{4}{5} + 1 - \frac{4}{7} + \frac{1}{8}$
$= \frac{1}{4} - \frac{4}{5} + 1 - \frac{4}{7} + \frac{1}{8}$
$= \frac{30}{120} - \frac{96}{120} + \frac{120}{120} - \frac{68.57}{120} + \frac{15}{120}$
$= \frac{30 - 96 + 120 - 68.57 + 15}{120}$
$= \frac{0.43}{120}$
$= \frac{1}{280}$ \\

\textbf{Denser (Final answer):} To find the value of $\int_0^1 x^3(1-x)^4 dx$, I'll use the binomial expansion of $(1-x)^4$ and integrate term by term.

First, I expand $(1-x)^4$ using the binomial theorem:
$(1-x)^4 = \sum_{k=0}^4 \binom{4}{k}(-1)^k x^k = 1 - 4x + 6x^2 - 4x^3 + x^4$

Multiplying by $x^3$:
$x^3(1-x)^4 = x^3 - 4x^4 + 6x^5 - 4x^6 + x^7$

Integrating term by term and evaluating from 0 to 1:
$\int_0^1 x^3(1-x)^4 dx = [\frac{x^4}{4} - \frac{4x^5}{5} + x^6 - \frac{4x^7}{7} + \frac{x^8}{8}]_0^1$
$= \frac{1}{4} - \frac{4}{5} + 1 - \frac{4}{7} + \frac{1}{8}$

Converting to a common denominator of 280:
$= \frac{70 - 224 + 280 - 160 + 35}{280} = \frac{1}{280}$

Therefore, $\int_0^1 x^3(1-x)^4 dx = \frac{1}{280}$.\\
\bottomrule
\end{tabular}}
\caption{Case study example. HD denotes the high-density information. Compared with solution of CoT, our Denser has more concise and notation-structured reasoning process.}
\label{tab:case_studies}
\end{table*}

\paragraph{(iii) Denser is robust to various reasoning types.} To understand where density optimization provides the most benefit, we analyze Denser's performance across different reasoning types. As shown in Table~\ref{tab:reasoning_types}, Denser demonstrates improvements for reasoning types with well-established symbolic notations like algebraic, geometric, and propositional logic. These domains benefit most from our approach because they possess mature, information-dense notational systems that efficiently encode complex relationships with minimal token usage. In contrast, domains where natural language already serves as the primary representational medium show more modest gains. Particularly noteworthy is the case of arithmetic reasoning, which despite being mathematically structured shows relatively modest improvements  compared to other mathematical domains, likely because standard arithmetic notation is already reasonably efficient and employed in CoT approaches. We also conduct analysis experiments on Reasoning Efficiency in Appendix~\ref{apd:efficiency}, Statistics Significant in Appendix~\ref{apd:statistics}, Various Model Sizes in Appendix~\ref{apd:various_models}.

\subsection{Case Studies and Qualitative Analysis}
\label{subsec:case_studies}

Table~\ref{tab:case_studies} presents detailed case study illustrating how Denser works in practice\footnote{More cases can be found in Appendix~\ref{apd:case study}.}. The CoT approach makes a calculation error when evaluating the final fraction. The high-density reasoning in Denser uses mathematical notation efficiently, making the solution path clearer and avoiding the arithmetic error. Notably, the HD representation uses symbolic expressions and systematic step-by-step evaluation that helps track intermediate values more precisely. The final Denser answer translates this compact reasoning into a more verbose but human-readable explanation while preserving accuracy.

\section{Conclusion}

In this paper, we introduced Denser, a novel dual-density inference framework that distinguishes between the computational and communicative functions of language in LLM reasoning. By implementing compressed, symbol-rich language for intermediate reasoning steps while maintaining human-readable final answers, our approach addresses a fundamental inefficiency in current LLM reasoning methods. Experimental results across multiple reasoning benchmarks demonstrate that Denser reduces token consumption by up to 62\% compared to Chain-of-Thought approaches while preserving or improving accuracy, with particular gains on complex multi-step problems.

\section*{Limitations}

Despite the effectiveness of our test-time scaling approach, several limitations remain. Our method introduces additional computational overhead at inference time, which may be prohibitive for latency-sensitive applications or deployment on resource-constrained devices. While we demonstrate improvements across several benchmarks, performance gains vary across different tasks and model architectures, suggesting that our approach may not uniformly benefit all NLP applications. Although computational demands present challenges, our method effectively addresses the fundamental trade-off between model performance and efficiency that plagues many NLP systems. Unlike approaches requiring complete model retraining or architectural modifications, our test-time scaling technique enables substantial performance improvements with existing models while avoiding the carbon footprint and resource costs of training new, larger models from scratch. Furthermore, while performance varies across tasks, our approach consistently improves results on complex reasoning and knowledge-intensive benchmarks where traditional scaling methods often plateau, demonstrating its value for advancing state-of-the-art capabilities in the most challenging NLP domains.

\bibliography{custom}

\appendix

\section{Scalability Discussion}
\label{apd:scalability_discussion}

A core motivation of \textbf{Denser} is the recognition that ``reasoning'' as a cognitive task possesses distinct structural properties that differentiate it from general open-ended generation. In this section, we discuss the scalability of symbolic representations compared to natural language and define the generalization boundaries of our framework.

\paragraph{The Scalability of Symbolic vs. Natural Language Representations.} 
Current LLMs predominantly utilize natural language (NL) as the medium for both internal deliberation and external communication. However, extensive research in formal logic and information theory suggests that NL is inherently suboptimal for scalable reasoning due to its low information density and high ambiguity. As problem complexity increases linearly, the token volume required to express the reasoning process in NL often grows super-linearly, leading to context window exhaustion and ``lost-in-the-middle'' phenomena \cite{liu2023lost}. 

Conversely, symbolic languages (mathematical notation, code, formal logic) are designed for \textit{compositional scalability}, they allow complex concepts to be nested and manipulated with minimal syntactic overhead. By shifting the internal reasoning mechanism to a high-density symbolic space, our framework effectively decouples the computational cost of reasoning from the communicative cost of explanation. This suggests that for tasks requiring deep inference steps (e.g., multi-step mathematical proofs or algorithmic planning), symbolic compression is not merely an optimization but a prerequisite for scaling to higher orders of complexity.

\paragraph{Generalization Across Structured Domains.} 
Regarding the scope of our method, we argue that the distinction should not be drawn between ``symbolic'' and ``non-symbolic'' tasks, but rather between \textit{structured reasoning} and \textit{semantic generation}. Critics might argue that a density-based approach is limited because it cannot be applied to creative writing or conversational nuances. We acknowledge this boundary and frame it as a necessary specialization. 

Reasoning tasks, whether in mathematics, coding, or logical deduction, share a common underlying characteristic: they are governed by rigid structural priors and deterministic state transitions. Our method generalizes across these domains because it exploits this shared structure. For instance, the ``elimination pattern'' in solving linear equations shares isomorphic properties with ``variable state tracking'' in code execution. Therefore, \textbf{Denser} is designed to generalize specifically within the domain of \textit{computational reasoning}. Attempting to apply high-density inference to tasks lacking this structure (e.g., literary analysis) would likely result in information loss, as the ``verbosity'' in those domains often carries essential semantic nuance. Thus, our work advocates for a specialized inference paradigm where the representation mode is dynamically aligned with the structural rigidity of the task.

\section{Further Extended Denser Framework}
\label{apd:extended method}

\subsection{Domain-Specific Adaptations}

Denser adapts its compression strategies to different reasoning domains based on their unique characteristics. For mathematical problems, we formalize the compression operator as:
\begin{equation}
C_{\text{math}}(r) = \sum_{j=1}^{n} \mathbb{I}(t_j \in \mathcal{T}_{\text{comp}}) \cdot \sigma_{\text{math}}(t_j)
\end{equation}
where $\mathcal{T}_{\text{comp}}$ represents computationally essential tokens and $\sigma_{\text{math}}$ maps verbose mathematical expressions to symbolic notation. This approach replaces expressions like ``To solve this equation, I'll first isolate the variable by subtracting 3 from both sides'' with concise notation like ``2x + 3 = 7 $\to$ 2x = 4 $\to$ x = 2'', achieving compression ratios of 3.2-4.5× for intermediate steps.

For logical reasoning, we implement:
\begin{equation}
C_{\text{logic}}(r) = \tau(\Lambda(r))
\end{equation}
where $\Lambda$ extracts logical propositions and $\tau$ converts them to symbolic form using propositional calculus notation ($\land$, $\lor$, $\lnot$, $\to$, $\leftrightarrow$) rather than words (``and,'' ``or,'' ``not'').

For coding and algorithmic problems, we employ:
\begin{equation}
C_{\text{code}}(r) = \rho(r, \mathcal{A})
\end{equation}
where $\rho$ maps verbose algorithm descriptions to pseudocode using algorithm templates $\mathcal{A}$. Variable assignments, loops, and conditionals use concise programming syntax rather than natural language descriptions.

The compression strategy for each domain is selected during query analysis and applied consistently throughout the reasoning process. Our approach automatically adjusts compression intensity based on problem complexity, applying more aggressive compression to more complex problems where the inefficiency of standard approaches is most pronounced.

\subsection{Density-Aware Prompting Strategy}

To implement Denser without model modifications, we develop density-aware prompts that guide language models to adopt appropriate information densities for different reasoning phases. Our prompting strategy is formalized as:
\begin{equation}
\mathcal{P}_{\text{HD}}(q) = \mathcal{P}_{\text{base}} \oplus \mathcal{E}(\mathcal{D}) \oplus \Delta(\kappa(q))
\end{equation}
where $\mathcal{P}_{\text{base}}$ contains core instructions, $\mathcal{E}(\mathcal{D})$ provides domain-specific exemplars, and $\Delta(\kappa(q))$ adjusts density parameters based on complexity. The framework employs dynamic adjustment of compression level:
\begin{equation}
\kappa(q) = \beta_0 + \beta_1 \cdot \eta(q)
\end{equation}
where $\eta(q)$ quantifies reasoning complexity through input features such as token length, presence of mathematical notation, and estimated reasoning steps. Parameters $\beta_0$ and $\beta_1$ control the base compression level and its scaling with complexity.

This information-theoretically motivated approach addresses the inefficiencies in standard reasoning methods, particularly for complex multi-step problems. By preserving the 32.4\% of tokens containing core computational content while dramatically reducing the 67.6\% that serve merely explanatory functions, Denser achieves significant efficiency gains while maintaining or improving reasoning accuracy. The framework naturally scales with problem complexity, applying stronger compression precisely when token efficiency matters most—a novel contribution that optimizes both computational and communicative aspects of language model reasoning.

\section{Prompts}
\label{apd:prompts}

This section provides the complete prompts used in our Denser framework. The prompts are organized by module and by domain specialization. Table~\ref{tab:query_analysis_prompt} to Table~\ref{tab:preliminary_exp_prompt} shows the prompts.

\begin{table*}[!h]
\centering
\small
\begin{tabular}{p{.9\linewidth}}
\toprule
\textbf{Query Analysis Prompt} \\
\midrule
You are an expert problem solver. I'll give you a problem. First, analyze what type of problem it is (mathematical, logical, computational, or general reasoning). Then, formulate a plan outlining the key steps needed to solve it.

For your analysis, provide:\\
1. Problem Type: [math/logic/code/general]\\
2. Complexity Estimate: [simple/moderate/complex]\\
3. Key Knowledge Areas Required: [list relevant areas]\\
4. Solution Approach: [brief description of approach]

Format your response as:\\
Problem Type: [type]\\
Complexity: [complexity]\\
Knowledge Areas: [areas]\\
Solution Plan:\\
- Step 1: [brief description]\\
- Step 2: [brief description]
...

Problem: [PROBLEM]\\
\bottomrule
\end{tabular}
\caption{Query Analysis Module Prompt}
\label{tab:query_analysis_prompt}
\end{table*}

\begin{table*}[!h]
\centering
\small
\begin{tabular}{p{.9\linewidth}}
\toprule
\textbf{High-density Mathematical Reasoning Prompt} \\
\midrule
You are a mathematical problem solver that uses compact symbolic notation and minimal natural language. Solve the problem step-by-step using high-density mathematical notation following these guidelines:

1. Use standard mathematical notation ($\sum$, $\int$, $\prod$, etc.) wherever possible\\
2. Skip obvious steps and explanatory text\\
3. Write equations directly without phrases like "I'll calculate" or "Now I compute"\\
4. Omit introductory and concluding sentences\\
5. Use compact algebraic expressions without expanding unless necessary\\
6. Include only the computational core of each step\\
7. For derivatives, integrals, and algebraic manipulations, show transformations directly\\
8. Use abbreviated notation for vectors, matrices, and sets\\
9. Present your solution as a sequence of expressions, each building on the previous

Problem: [PROBLEM]\\
Solution Plan: [SOLUTION PLAN]

Provide your solution with maximum mathematical density:
\\
\bottomrule
\end{tabular}
\caption{High-density Mathematical Reasoning Prompt}
\label{tab:high_density_math}
\end{table*}

\begin{table*}[!h]
\centering
\small
\begin{tabular}{p{.9\linewidth}}
\toprule
\textbf{High-density Logical Reasoning Prompt} \\
\midrule
You are a logical problem solver that uses propositional calculus and symbolic notation with minimal natural language. Solve the problem step-by-step using high-density logical notation following these guidelines:

1. Use logical symbols ($\land$, $\lor$, $\lnot$, $\to$, $\leftrightarrow$, $\forall$, $\exists$) rather than words\\
2. Represent premises and conclusions with single letters or clear abbreviations
3. Skip explanatory text and justifications\\
4. Present each inference rule application directly\\
5. Use symbols for logical rules (MP for Modus Ponens, MT for Modus Tollens)\\
6. Minimize parenthetical text and verbose explanations\\
7. Represent arguments in formal proof notation with numbered lines\\
8. For complex logical arguments, use truth tables or proof trees without explanation\\
9. Present only the essential logical structure of each deduction

Problem: [PROBLEM]\\
Solution Plan: [SOLUTION PLAN]

Provide your solution with maximum logical density:
\\
\bottomrule
\end{tabular}
\caption{High-density Logical Reasoning Prompt}
\label{tab:high_density_logic}
\end{table*}

\begin{table*}[!h]
\centering
\small
\begin{tabular}{p{.9\linewidth}}
\toprule
\textbf{High-density Computational Reasoning Prompt} \\
\midrule
You are a computational problem solver that uses compact pseudocode and algorithmic notation with minimal natural language. Solve the problem step-by-step using high-density computational notation following these guidelines:

1. Use pseudocode rather than fully implemented code\\
2. Skip comments and explanatory text\\
3. Use standard algorithmic notation (Big O, loops, conditionals)\\
4. Minimize variable declarations and boilerplate\\
5. Use mathematical notation for array/set operations\\
6. Employ standard abbreviations (arr for array, idx for index)\\
7. Skip obvious implementation details\\
8. Use compact control flow representation\\
9. Represent data structures schematically rather than with detailed implementations

Problem: [PROBLEM]\\
Solution Plan: [SOLUTION PLAN]

Provide your solution with maximum computational density:
\\
\bottomrule
\end{tabular}
\caption{High-density Computational Reasoning Prompt}
\label{tab:high_density_comp}
\end{table*}

\begin{table*}[!h]
\centering
\small
\begin{tabular}{p{.9\linewidth}}
\toprule
\textbf{High-density General Reasoning Prompt} \\
\midrule
You are a problem solver that uses compact, telegraphic language with minimal verbosity. Solve the problem step-by-step using high-density reasoning following these guidelines:

1. Use brief, telegraphic sentences without articles or connecting words\\
2. Skip obvious steps and explanatory text\\
3. Use abbreviations where meaning is clear\\
4. Present reasoning as bullet points or numbered steps\\
5. Use symbolic notation where applicable\\
6. Minimize repetition\\
7. Skip introductory and concluding statements\\
8. Focus only on essential information for solving\\
9. Structure response as sequence of direct operations

Problem: [PROBLEM]\\
Solution Plan: [SOLUTION PLAN]

Provide your solution with maximum reasoning density:
\\
\bottomrule
\end{tabular}
\caption{High-density General Reasoning Prompt}
\label{tab:high_density_general}
\end{table*}

\begin{table*}[!h]
\centering
\small
\begin{tabular}{p{.9\linewidth}}
\toprule
\textbf{Low-density Answering Module Prompt} \\
\midrule
You are an expert problem solver tasked with creating a human-readable explanation of a solution. I'll provide a problem, a plan, and a high-density reasoning trace. Your task is to convert this into a clear, well-explained answer following these guidelines:

1. Begin with a brief restatement of the approach\\
2. Explain key steps from the reasoning trace in natural language\\
3. Include all important mathematical operations, but explain them clearly\\
4. Connect steps with logical transitions\\
5. Highlight critical insights and turning points in the solution\\
6. Expand abbreviated notation into clear explanations\\
7. Conclude with a clearly stated final answer\\
8. Ensure the explanation would be understandable to someone with basic knowledge of the field

Problem: [PROBLEM]\\
Solution Plan: [SOLUTION PLAN]\\
High-density Reasoning: [HD REASONING]

Provide a well-explained, human-readable solution:
\\
\bottomrule
\end{tabular}
\caption{Low-density Answering Module Prompt}
\label{tab:low_density_answering}
\end{table*}

\begin{table*}[!h]
\centering
\small
\begin{tabular}{p{.9\linewidth}}
\toprule
\textbf{Mathematical Low-density Answer Extensions} \\
\midrule
Additional guidance for mathematical explanation:

1. Explain the mathematical intuition behind key steps\\
2. Present equations in a readable format with clear explanation of variables\\
3. Provide rationale for using specific techniques or formulas\\
4. Highlight connections to mathematical principles\\
5. Explain any non-obvious algebraic manipulations\\
6. Show complete calculations for critical steps\\
7. Structure the explanation to follow the logical progression of the solution

Final answer should include:\\
- Clear statement of the numerical or symbolic result\\
- Units if applicable\\
- Verification of solution if possible
\\
\bottomrule
\end{tabular}
\caption{Mathematical Low-density Answer Extensions}
\label{tab:math_low_density_ext}
\end{table*}

\begin{table*}[!h]
\centering
\small
\begin{tabular}{p{.9\linewidth}}
\toprule
\textbf{Logical Low-density Answer Extensions} \\
\midrule
Additional guidance for logical explanation:

1. Explain the logical principles applied at each key step\\
2. Translate symbolic representation into natural language\\
3. Clarify the relationship between premises and conclusions\\
4. Explain the application of logical rules and theorems\\
5. Highlight the structure of the logical argument\\
6. Connect the logical reasoning to real-world implications where relevant

Final answer should include:\\
- Clear statement of the conclusion\\
- Assessment of argument validity\\
- Discussion of any limitations or assumptions
\\
\bottomrule
\end{tabular}
\caption{Logical Low-density Answer Extensions}
\label{tab:logic_low_density_ext}
\end{table*}

\begin{table*}[!h]
\centering
\small
\begin{tabular}{p{.9\linewidth}}
\toprule
\textbf{Computational Low-density Answer Extensions} \\
\midrule
Additional guidance for computational explanation:

1. Explain the algorithmic approach and its rationale\\
2. Clarify the function of each major code block or operation\\
3. Discuss time and space complexity considerations\\
4. Explain data structure choices and tradeoffs\\
5. Highlight key algorithmic insights\\
6. Use concrete examples to illustrate algorithm behavior\\
7. Explain how edge cases are handled

Final answer should include:\\
- Complete algorithmic solution\\
- Runtime and space complexity analysis\\
- Discussion of alternative approaches if relevant
\\
\bottomrule
\end{tabular}
\caption{Computational Low-density Answer Extensions}
\label{tab:comp_low_density_ext}
\end{table*}

\begin{table*}[!h]
\centering
\small
\begin{tabular}{p{\linewidth}}
\toprule
\textbf{Preliminary Experiment: Compression-Oriented Prompt} \\
\midrule
You are an expert problem solver who uses compact, information-dense language for reasoning. When solving problems, follow these guidelines:

1. Use symbolic notation instead of verbose descriptions\\
2. Skip obvious steps that don't require explanation\\
3. Use mathematical expressions and equations directly\\
4. Minimize explanatory phrases like "I'll now calculate" or "Next, I need to find"\\
5. Present each step as concisely as possible while maintaining clarity\\
6. Use standard abbreviations and symbols when appropriate\\
7. Focus on the computational core of each reasoning step

For example, instead of writing:\\
"To find the derivative of f(x) = x$^2$, I'll use the power rule which states that the derivative of x$^n$ is nx$^{(n-1)}$. Applying this rule to x$^2$, where n = 2, I get 2x$^1$, which simplifies to 2x."

Write:\\
"f(x) = x$^2$\\
f$^\prime$(x) = 2x"

Solve the following problem step-by-step using this compressed reasoning style: [PROBLEM]\\
\bottomrule
\end{tabular}
\caption{Preliminary Experiment Compression-Oriented Prompt}
\label{tab:preliminary_exp_prompt}
\end{table*}

\section{Detailed Experimental Settings}
\label{appendix:experimental_details}

\subsection{Dataset Specifications and Preprocessing}

Our experimental evaluation spans eight carefully selected datasets that represent diverse reasoning challenges across four primary domains. For mathematical reasoning, we utilize GSM8K, which contains 8,500 grade school math word problems requiring multi-step arithmetic reasoning, and MATH500, a subset of the original MATH dataset comprising 500 competition-level mathematics problems spanning algebra, geometry, number theory, and calculus. Each problem in GSM8K requires an average of 3.2 reasoning steps, while MATH500 problems demand significantly more complex reasoning with an average of 7.8 steps per solution.

In the logical reasoning domain, we employ LogiQA, consisting of 651 multiple-choice questions that test logical reasoning abilities through reading comprehension scenarios, and ProofWriter, which contains 5,000 problems requiring formal logical deduction from given premises. LogiQA problems typically involve identifying logical relationships, drawing inferences, and evaluating argument validity, while ProofWriter focuses on multi-hop reasoning through logical rules and fact verification.

For code reasoning evaluation, we select MBPP (Mostly Basic Python Problems), containing 974 programming tasks that require algorithm design and implementation, and HumanEval, which provides 164 function-level programming challenges with detailed docstring specifications. Both datasets evaluate the model's ability to translate natural language problem descriptions into executable code, with MBPP focusing on basic programming constructs and HumanEval emphasizing more sophisticated algorithmic thinking.

The general reasoning category incorporates MMLU (Massive Multitask Language Understanding), from which we sample 1,000 questions across STEM subjects, humanities, and social sciences that require factual knowledge integration with reasoning, and StrategyQA, containing 2,290 questions that demand multi-hop reasoning and implicit strategy formulation to reach correct conclusions.

Data preprocessing follows consistent protocols across all datasets. We standardize question formats to ensure compatibility with our framework components, remove any examples with incomplete or ambiguous ground truth annotations, and apply minimal text normalization while preserving mathematical notation and code structure. For datasets with multiple-choice formats, we convert them to generation tasks to better evaluate reasoning quality, and for coding tasks, we establish execution environments with appropriate timeout and memory constraints.

\subsection{Model Configurations and Implementation Details}

Our primary experimental backbone utilizes Qwen3-14B, chosen for its strong performance across diverse reasoning tasks and its accessibility for controlled experimentation. We configure the model with a temperature of 0.7 for generation diversity while maintaining reproducibility through fixed random seeds. The maximum sequence length is set to 4,096 tokens for input and 2,048 tokens for output generation, with early stopping triggered when the model generates explicit termination tokens.

For API-based model evaluation, we employ consistent parameter settings across providers to ensure fair comparison. GPT-4o and o1-mini experiments use the default API parameters with temperature set to 0.7 and max tokens configured based on task requirements. DeepSeek-V3 and DeepSeek-R1 evaluations follow similar configurations, with special attention to their reasoning-optimized architectures. Claude-3.7-Sonnet experiments leverage the thinking mode when available, allowing direct comparison with our dual-density approach.

The implementation of our Denser framework involves carefully designed prompt templates for each reasoning domain. Mathematical reasoning prompts incorporate LaTeX notation support and algebraic manipulation shortcuts. Logical reasoning templates emphasize propositional logic notation and truth table representations. Code reasoning prompts encourage pseudocode development and algorithmic abstraction. General reasoning templates focus on structured argumentation with minimal verbosity.

Hyperparameter optimization for domain-specific components follows a grid search methodology over compression intensity parameters $\kappa_{\mathcal{D}} \in \{0.3, 0.5, 0.7\}$, information preservation thresholds $\theta \in \{0.90, 0.95, 0.99\}$, and importance weighting schemes for answer generation. We select optimal configurations through validation set performance on a held-out subset of 100 examples per dataset.

\subsection{Baseline Implementation and Evaluation Protocols}

We implement seven established baseline methods with careful attention to fair comparison principles. Chain-of-Thought prompting follows the original formulation with ``Let's think step by step'' instructions and detailed intermediate reasoning generation. Self-Consistency employs five independent reasoning paths with majority voting for final answer selection. Tree-of-Thought implementation uses breadth-first search with three candidate paths per reasoning step and evaluation-based pruning.

Think-to-Think baseline follows the original paper's implementation with recursive thinking patterns and meta-cognitive reflection. Reflection methodology incorporates self-critique and answer refinement through iterative improvement cycles. Process Supervision integrates step-wise verification with intermediate correctness checking. Self-Verification implements answer validation through independent verification pathways.

All baseline implementations use identical underlying models and generation parameters to ensure controlled comparison. We standardize the evaluation environment through consistent hardware configurations and maintaining stable network conditions for API-based assessments.

Evaluation protocols follow rigorous statistical testing procedures. We compute performance metrics across three independent runs with different random seeds and report mean values with standard deviations. Accuracy evaluation uses exact match for mathematical problems, execution success for coding tasks, and semantic equivalence for logical reasoning. Token counting includes both input prompt tokens and generated output tokens, with consistent tokenization using the respective model's native tokenizer.

Statistical significance testing employs paired t-tests with Bonferroni correction for multiple comparisons across datasets and methods. We establish significance thresholds at $p < 0.05$ after correction and report confidence intervals for all primary metrics. Runtime measurements exclude model loading time and focus on actual inference computation, averaged across multiple runs to account for system variability.

The Reasoning Efficiency Index (REI) computation normalizes token usage relative to Chain-of-Thought baseline performance, providing a unified metric that balances accuracy improvements against computational overhead. This metric enables direct comparison across different reasoning approaches and highlights methods that achieve superior efficiency-accuracy trade-offs in practical deployment scenarios.

\section{Comprehensive Experimental Results}
\label{apd:comprehensive_results}

This section provides detailed experimental results across multiple model architectures, sizes, and configurations, demonstrating the broad applicability and consistent effectiveness of our dual-density inference framework. We evaluate Denser against seven baseline methods across four reasoning domains to establish comprehensive performance benchmarks.

\subsection{Complete Baseline Comparison Across Model Variants}

Table~\ref{tab:comprehensive_baselines} presents exhaustive comparisons between Denser and all baseline methods across the Qwen3 model family. The results demonstrate that Denser consistently outperforms all existing approaches while maintaining substantial token efficiency gains.

\begin{table*}[!t]
\centering
\resizebox{\textwidth}{!}{
\begin{tabular}{llcccccccc}
\toprule
\multirow{3}{*}{\textbf{Model}} & \multirow{3}{*}{\textbf{Method}} & \multicolumn{8}{c}{\textbf{Performance Accuracy (\%)}} \\
\cmidrule(lr){3-10}
& & \multicolumn{2}{c}{\textbf{Math}} & \multicolumn{2}{c}{\textbf{Logic}} & \multicolumn{2}{c}{\textbf{Code}} & \multicolumn{2}{c}{\textbf{General}} \\
\cmidrule(lr){3-4} \cmidrule(lr){5-6} \cmidrule(lr){7-8} \cmidrule(lr){9-10}
& & GSM8K & MATH & LogiQA & ProofW & MBPP & HumanE & MMLU & StratQA \\
\midrule
\multirow{8}{*}{Qwen3-4B} & Chain-of-Thought & 76.2 & 32.4 & 61.7 & 65.8 & 54.3 & 68.2 & 72.1 & 69.5 \\
& Self-Consistency & 78.1 & 34.9 & 63.8 & 67.4 & 56.7 & 70.5 & 74.3 & 71.8 \\
& Tree-of-Thought & 78.6 & 35.4 & 64.2 & 68.1 & 57.2 & 71.0 & 74.7 & 72.3 \\
& Think-to-Think & 77.5 & 33.8 & 63.1 & 66.9 & 55.8 & 69.7 & 73.5 & 70.9 \\
& Reflection-CoT & 79.3 & 36.1 & 65.5 & 69.2 & 58.4 & 72.1 & 75.8 & 73.4 \\
& Process Supervision & 79.8 & 36.7 & 66.0 & 69.8 & 59.1 & 72.7 & 76.2 & 74.0 \\
& Self-Verification & 79.1 & 35.6 & 64.9 & 68.4 & 57.8 & 71.3 & 75.2 & 72.8 \\
& \textbf{Denser (Ours)} & \textbf{80.7} & \textbf{37.2} & \textbf{66.8} & \textbf{70.1} & \textbf{59.4} & \textbf{73.6} & \textbf{76.9} & \textbf{74.6} \\
\midrule
\multirow{8}{*}{Qwen3-4B-think} & Chain-of-Thought & 78.9 & 34.7 & 63.5 & 67.9 & 56.8 & 70.4 & 74.3 & 71.8 \\
& Self-Consistency & 80.8 & 37.2 & 65.9 & 69.7 & 59.1 & 72.8 & 76.5 & 74.2 \\
& Tree-of-Thought & 81.3 & 37.8 & 66.4 & 70.3 & 59.7 & 73.4 & 77.0 & 74.8 \\
& Think-to-Think & 80.1 & 36.5 & 65.2 & 69.0 & 58.4 & 72.1 & 75.8 & 73.5 \\
& Reflection-CoT & 82.0 & 38.5 & 67.8 & 71.5 & 60.9 & 74.6 & 78.2 & 76.1 \\
& Process Supervision & 82.4 & 39.1 & 68.3 & 72.1 & 61.5 & 75.2 & 78.6 & 76.7 \\
& Self-Verification & 81.6 & 38.2 & 67.1 & 71.0 & 60.3 & 73.7 & 77.5 & 75.2 \\
& \textbf{Denser (Ours)} & \textbf{83.4} & \textbf{40.1} & \textbf{69.4} & \textbf{73.2} & \textbf{62.5} & \textbf{76.1} & \textbf{79.1} & \textbf{77.4} \\
\midrule
\multirow{8}{*}{Qwen3-8B} & Chain-of-Thought & 80.4 & 37.8 & 65.2 & 69.3 & 58.7 & 71.5 & 75.8 & 73.1 \\
& Self-Consistency & 82.3 & 40.5 & 67.6 & 71.5 & 61.2 & 73.9 & 77.9 & 75.8 \\
& Tree-of-Thought & 82.8 & 41.1 & 68.1 & 72.2 & 61.8 & 74.5 & 78.4 & 76.4 \\
& Think-to-Think & 81.7 & 39.2 & 66.9 & 70.7 & 60.5 & 73.2 & 77.2 & 75.1 \\
& Reflection-CoT & 83.9 & 42.0 & 69.4 & 73.4 & 63.1 & 75.7 & 79.6 & 77.6 \\
& Process Supervision & 84.3 & 42.6 & 69.9 & 74.0 & 63.7 & 76.3 & 80.0 & 78.2 \\
& Self-Verification & 83.7 & 41.5 & 68.8 & 72.6 & 62.4 & 74.8 & 78.6 & 76.5 \\
& \textbf{Denser (Ours)} & \textbf{85.1} & \textbf{43.2} & \textbf{70.9} & \textbf{74.8} & \textbf{64.7} & \textbf{77.3} & \textbf{80.2} & \textbf{78.9} \\
\midrule
\multirow{8}{*}{Qwen3-8B-think} & Chain-of-Thought & 82.1 & 39.6 & 66.9 & 71.2 & 60.5 & 73.8 & 77.4 & 74.7 \\
& Self-Consistency & 84.2 & 42.4 & 69.4 & 73.7 & 63.1 & 76.2 & 79.6 & 77.5 \\
& Tree-of-Thought & 84.7 & 43.0 & 69.9 & 74.4 & 63.7 & 76.8 & 80.1 & 78.1 \\
& Think-to-Think & 83.5 & 41.1 & 68.7 & 72.9 & 62.4 & 75.5 & 78.9 & 76.8 \\
& Reflection-CoT & 85.8 & 44.3 & 71.2 & 75.6 & 65.0 & 77.9 & 81.3 & 79.3 \\
& Process Supervision & 86.2 & 44.9 & 71.7 & 76.2 & 65.6 & 78.5 & 81.7 & 79.9 \\
& Self-Verification & 85.3 & 43.9 & 70.5 & 74.9 & 64.2 & 76.9 & 80.1 & 78.3 \\
& \textbf{Denser (Ours)} & \textbf{86.8} & \textbf{45.7} & \textbf{72.6} & \textbf{76.7} & \textbf{66.8} & \textbf{79.1} & \textbf{81.8} & \textbf{80.1} \\
\midrule
\multirow{8}{*}{Qwen3-14B-no-think} & Chain-of-Thought & 81.9 & 39.5 & 66.7 & 70.4 & 59.8 & 71.2 & 75.9 & 73.8 \\
& Self-Consistency & 83.8 & 42.3 & 69.2 & 72.9 & 62.4 & 73.6 & 78.1 & 76.5 \\
& Tree-of-Thought & 84.3 & 42.9 & 69.7 & 73.6 & 63.0 & 74.2 & 78.6 & 77.1 \\
& Think-to-Think & 83.2 & 41.0 & 68.5 & 72.1 & 61.7 & 72.9 & 77.4 & 75.8 \\
& Reflection-CoT & 85.4 & 43.8 & 70.9 & 74.8 & 64.3 & 75.4 & 79.8 & 78.3 \\
& Process Supervision & 85.8 & 44.4 & 71.4 & 75.4 & 64.9 & 76.0 & 80.2 & 78.9 \\
& Self-Verification & 84.6 & 44.1 & 70.3 & 73.9 & 63.7 & 74.6 & 78.7 & 77.1 \\
& \textbf{Denser (Ours)} & \textbf{86.4} & \textbf{45.8} & \textbf{72.5} & \textbf{75.8} & \textbf{65.9} & \textbf{76.8} & \textbf{80.3} & \textbf{78.9} \\
\midrule
\multirow{8}{*}{Qwen3-32B} & Chain-of-Thought & 87.3 & 48.7 & 71.8 & 76.4 & 66.2 & 77.8 & 81.5 & 79.6 \\
& Self-Consistency & 89.4 & 51.9 & 74.5 & 79.2 & 69.1 & 80.3 & 83.8 & 82.4 \\
& Tree-of-Thought & 90.0 & 52.6 & 75.1 & 79.9 & 69.8 & 80.9 & 84.3 & 83.0 \\
& Think-to-Think & 88.7 & 50.2 & 73.8 & 78.5 & 68.4 & 79.6 & 83.1 & 81.7 \\
& Reflection-CoT & 90.8 & 53.9 & 76.3 & 81.1 & 71.1 & 82.1 & 85.5 & 84.2 \\
& Process Supervision & 91.2 & 54.5 & 76.8 & 81.7 & 71.7 & 82.7 & 85.9 & 84.8 \\
& Self-Verification & 90.2 & 53.4 & 75.6 & 80.1 & 70.8 & 81.4 & 84.7 & 83.2 \\
& \textbf{Denser (Ours)} & \textbf{91.7} & \textbf{55.1} & \textbf{77.3} & \textbf{82.0} & \textbf{72.5} & \textbf{83.6} & \textbf{86.2} & \textbf{85.1} \\
\midrule
\multirow{8}{*}{Qwen3-32B-think} & Chain-of-Thought & 89.1 & 50.9 & 73.5 & 78.2 & 68.4 & 79.7 & 83.2 & 81.4 \\
& Self-Consistency & 91.4 & 54.3 & 76.3 & 81.1 & 71.2 & 82.4 & 85.6 & 84.2 \\
& Tree-of-Thought & 92.0 & 55.0 & 76.9 & 81.8 & 71.9 & 83.0 & 86.1 & 84.8 \\
& Think-to-Think & 90.7 & 52.6 & 75.6 & 80.3 & 70.5 & 81.7 & 84.9 & 83.5 \\
& Reflection-CoT & 92.8 & 56.3 & 78.1 & 82.9 & 73.2 & 84.2 & 87.3 & 86.0 \\
& Process Supervision & 93.2 & 56.9 & 78.6 & 83.5 & 73.8 & 84.8 & 87.7 & 86.6 \\
& Self-Verification & 92.0 & 55.8 & 77.2 & 81.9 & 72.6 & 83.1 & 86.4 & 84.8 \\
& \textbf{Denser (Ours)} & \textbf{93.4} & \textbf{57.6} & \textbf{78.9} & \textbf{83.7} & \textbf{74.3} & \textbf{85.2} & \textbf{87.8} & \textbf{86.5} \\
\bottomrule
\end{tabular}}
\caption{Comprehensive baseline comparison across Qwen3 model family.}
\label{tab:comprehensive_baselines}
\end{table*}

\subsection{Token Efficiency Analysis with Detailed Breakdown}

Table~\ref{tab:token_efficiency_breakdown} provides granular token consumption analysis across all baselines and model configurations, revealing the substantial computational advantages of dual-density inference.

\begin{table*}[!t]
\centering
\resizebox{\textwidth}{!}{
\begin{tabular}{llccccccc}
\toprule
\multirow{2}{*}{\textbf{Model}} & \multirow{2}{*}{\textbf{Method}} & \multicolumn{4}{c}{\textbf{Avg. Tokens per Dataset}} & \multirow{2}{*}{\textbf{Overall Avg.}} & \multirow{2}{*}{\textbf{Reduction (\%)}} & \multirow{2}{*}{\textbf{REI Score}} \\
\cmidrule(lr){3-6}
 & & GSM8K & LogiQA & MBPP & MMLU & & & \\
\midrule
\multirow{8}{*}{Qwen3-4B} & Chain-of-Thought & 3,124 & 2,456 & 3,189 & 2,619 & 2,847 & - & - \\
 & Self-Consistency & 9,341 & 7,364 & 9,567 & 7,857 & 8,532 & +199.6 & -2.18 \\
 & Tree-of-Thought & 10,782 & 8,519 & 11,056 & 9,085 & 9,861 & +246.4 & -2.44 \\
 & Think-to-Think & 5,982 & 4,731 & 6,143 & 5,050 & 5,477 & +92.4 & -1.31 \\
 & Reflection-CoT & 7,124 & 5,632 & 7,314 & 6,012 & 6,521 & +129.1 & -1.46 \\
 & Process Supervision & 7,891 & 6,241 & 8,103 & 6,666 & 7,225 & +153.8 & -1.52 \\
 & Self-Verification & 7,341 & 5,804 & 7,538 & 6,201 & 6,721 & +136.1 & -1.51 \\
 & \textbf{Denser (Ours)} & \textbf{1,468} & \textbf{1,203} & \textbf{1,389} & \textbf{1,254} & \textbf{1,329} & \textbf{-53.3} & \textbf{0.48} \\
\midrule
\multirow{8}{*}{Qwen3-14B} & Chain-of-Thought & 4,021 & 3,487 & 4,234 & 3,609 & 3,838 & - & - \\
 & Self-Consistency & 12,063 & 10,461 & 12,702 & 10,827 & 11,513 & +200.1 & -1.98 \\
 & Tree-of-Thought & 13,925 & 12,080 & 14,672 & 12,510 & 13,297 & +246.5 & -2.34 \\
 & Think-to-Think & 7,519 & 6,523 & 7,918 & 6,753 & 7,178 & +87.0 & -1.19 \\
 & Reflection-CoT & 8,730 & 7,574 & 9,203 & 7,842 & 8,337 & +117.3 & -1.32 \\
 & Process Supervision & 9,652 & 8,373 & 10,168 & 8,670 & 9,216 & +140.2 & -1.41 \\
 & Self-Verification & 9,053 & 7,854 & 9,541 & 8,135 & 8,646 & +125.3 & -1.38 \\
 & \textbf{Denser (Ours)} & \textbf{1,591} & \textbf{1,347} & \textbf{1,653} & \textbf{1,425} & \textbf{1,504} & \textbf{-60.8} & \textbf{0.82} \\
\midrule
\multirow{8}{*}{Qwen3-32B} & Chain-of-Thought & 4,347 & 3,821 & 4,562 & 3,926 & 4,164 & - & - \\
 & Self-Consistency & 13,041 & 11,463 & 13,686 & 11,778 & 12,492 & +199.8 & -1.87 \\
 & Tree-of-Thought & 15,054 & 13,226 & 15,795 & 13,596 & 14,418 & +246.3 & -2.21 \\
 & Think-to-Think & 8,127 & 7,143 & 8,533 & 7,342 & 7,786 & +87.0 & -1.07 \\
 & Reflection-CoT & 9,441 & 8,298 & 9,912 & 8,533 & 9,046 & +117.3 & -1.19 \\
 & Process Supervision & 10,437 & 9,171 & 10,959 & 9,433 & 10,000 & +140.2 & -1.27 \\
 & Self-Verification & 9,797 & 8,608 & 10,283 & 8,848 & 9,384 & +125.3 & -1.24 \\
 & \textbf{Denser (Ours)} & \textbf{1,521} & \textbf{1,298} & \textbf{1,639} & \textbf{1,376} & \textbf{1,459} & \textbf{-65.0} & \textbf{0.95} \\
\bottomrule
\end{tabular}}
\caption{Detailed token efficiency breakdown across datasets and model sizes.}
\label{tab:token_efficiency_breakdown}
\end{table*}

\subsection{Cross-Architecture Validation on API Models}

To establish broader generalizability beyond the Qwen3 family, Table~\ref{tab:api_comprehensive} presents results across diverse API-based models, including reasoning-specialized architectures.

\begin{table*}[!t]
\centering
\resizebox{\textwidth}{!}{
\begin{tabular}{llcccccccc}
\toprule
\multirow{3}{*}{\textbf{Model}} & \multirow{3}{*}{\textbf{Method}} & \multicolumn{8}{c}{\textbf{Performance Accuracy (\%)}} \\
\cmidrule(lr){3-10}
& & \multicolumn{2}{c}{\textbf{Math}} & \multicolumn{2}{c}{\textbf{Logic}} & \multicolumn{2}{c}{\textbf{Code}} & \multicolumn{2}{c}{\textbf{General}} \\
\cmidrule(lr){3-4} \cmidrule(lr){5-6} \cmidrule(lr){7-8} \cmidrule(lr){9-10}
& & GSM8K & MATH & LogiQA & ProofW & MBPP & HumanE & MMLU & StratQA \\
\midrule
\multirow{8}{*}{GPT-4o} & Chain-of-Thought & 89.7 & 52.4 & 74.6 & 78.9 & 68.5 & 79.8 & 85.7 & 83.1 \\
& Self-Consistency & 91.8 & 55.7 & 77.1 & 81.6 & 71.2 & 82.5 & 87.4 & 85.8 \\
& Tree-of-Thought & 92.4 & 56.4 & 77.7 & 82.3 & 71.9 & 83.2 & 87.9 & 86.4 \\
& Think-to-Think & 91.2 & 54.1 & 76.4 & 80.9 & 70.5 & 81.8 & 86.7 & 85.1 \\
& Reflection-CoT & 92.9 & 57.2 & 78.4 & 83.0 & 72.6 & 84.0 & 88.6 & 87.2 \\
& Process Supervision & 93.3 & 57.8 & 78.9 & 83.6 & 73.2 & 84.6 & 89.0 & 87.8 \\
& Self-Verification & 92.3 & 56.9 & 77.8 & 82.4 & 72.6 & 83.7 & 87.9 & 86.4 \\
& \textbf{Denser (Ours)} & \textbf{93.5} & \textbf{58.2} & \textbf{79.1} & \textbf{83.5} & \textbf{73.8} & \textbf{84.9} & \textbf{88.7} & \textbf{87.1} \\
\midrule
\multirow{8}{*}{o1-mini} & Chain-of-Thought & 85.8 & 44.6 & 70.2 & 75.1 & 64.7 & 75.8 & 80.2 & 77.9 \\
& Self-Consistency & 87.9 & 47.8 & 72.8 & 77.9 & 67.1 & 78.2 & 82.1 & 80.4 \\
& Tree-of-Thought & 88.5 & 48.5 & 73.4 & 78.6 & 67.8 & 78.9 & 82.6 & 81.0 \\
& Think-to-Think & 87.2 & 46.1 & 71.9 & 77.2 & 66.4 & 77.5 & 81.4 & 79.7 \\
& Reflection-CoT & 89.2 & 49.6 & 74.8 & 79.3 & 68.9 & 80.1 & 83.9 & 82.3 \\
& Process Supervision & 89.6 & 50.2 & 75.3 & 79.9 & 69.5 & 80.7 & 84.3 & 82.9 \\
& Self-Verification & 88.9 & 49.2 & 74.3 & 78.7 & 68.3 & 79.5 & 82.8 & 81.6 \\
& \textbf{Denser (Ours)} & \textbf{90.4} & \textbf{50.7} & \textbf{75.8} & \textbf{80.1} & \textbf{69.5} & \textbf{80.8} & \textbf{84.0} & \textbf{82.7} \\
\midrule
\multirow{8}{*}{DeepSeek-V3} & Chain-of-Thought & 88.2 & 50.1 & 72.5 & 77.4 & 66.9 & 78.2 & 83.8 & 81.5 \\
& Self-Consistency & 90.1 & 53.2 & 75.0 & 80.1 & 69.6 & 80.9 & 85.7 & 84.2 \\
& Tree-of-Thought & 90.7 & 53.9 & 75.6 & 80.8 & 70.3 & 81.6 & 86.2 & 84.8 \\
& Think-to-Think & 89.4 & 51.5 & 74.3 & 79.4 & 68.9 & 80.2 & 85.0 & 83.5 \\
& Reflection-CoT & 91.4 & 55.0 & 76.9 & 81.5 & 71.0 & 82.3 & 87.4 & 85.9 \\
& Process Supervision & 91.8 & 55.6 & 77.4 & 82.1 & 71.6 & 82.9 & 87.8 & 86.5 \\
& Self-Verification & 91.0 & 54.3 & 75.6 & 80.9 & 70.4 & 81.9 & 86.2 & 84.7 \\
& \textbf{Denser (Ours)} & \textbf{92.1} & \textbf{55.8} & \textbf{76.9} & \textbf{82.0} & \textbf{71.7} & \textbf{83.1} & \textbf{87.1} & \textbf{85.6} \\
\bottomrule
\end{tabular}}
\caption{Comprehensive baseline comparison on API-based models.}
\label{tab:api_comprehensive}
\end{table*}

\section{Extended Analysis Experiments}
\label{apd:extend analysis}

\subsection{Analysis on Reasoning Efficiency Index}
\label{apd:efficiency}

Table~\ref{tab:efficiency} provides a detailed efficiency analysis using the Reasoning Efficiency Index (REI). The REI score rigorously quantifies the trade-off between accuracy and token consumption. Traditional TTS methods, while increasing accuracy, have heavily negative REI scores due to their substantial token overhead. In contrast, Denser achieves the highest accuracy with significantly fewer tokens, resulting in positive and high REI scores across all tested domains.

\begin{table}[!t]
\centering
\resizebox{\columnwidth}{!}{
\begin{tabular}{lccc}
\toprule
\textbf{Method} & \textbf{Accuracy (\%)} & \textbf{Total Tokens} & \textbf{REI Score} \\
\midrule
\multicolumn{4}{l}{\textit{GSM8K (Math)}} \\
\midrule
CoT & 83.6 & 3,842 & - \\
Reflection-CoT & 86.1 & 8,738 & -1.27 \\
Self-Verification & 87.0 & 9,304 & -1.42 \\
\textbf{Denser (Ours)} & \textbf{88.2} & \textbf{1,587} & \textbf{0.79} \\
\midrule
\multicolumn{4}{l}{\textit{LogiQA (Logic)}} \\
\midrule
CoT & 68.4 & 2,973 & - \\
Reflection-CoT & 71.8 & 6,794 & -0.64 \\
Self-Verification & 72.9 & 7,213 & -0.75 \\
\textbf{Denser (Ours)} & \textbf{74.1} & \textbf{1,356} & \textbf{0.64} \\
\midrule
\multicolumn{4}{l}{\textit{MBPP (Code)}} \\
\midrule
CoT & 61.8 & 4,217 & - \\
Reflection-CoT & 64.9 & 9,341 & -0.81 \\
Self-Verification & 66.1 & 10,175 & -0.92 \\
\textbf{Denser (Ours)} & \textbf{67.3} & \textbf{1,703} & \textbf{0.60} \\
\bottomrule
\end{tabular}}
\caption{Detailed efficiency analysis on representative datasets. Denser achieves REI scores of 0.60-0.79 across domains, demonstrating the highest accuracy-efficiency trade-off.}
\label{tab:efficiency}
\end{table}

\subsection{Statistical Significance Analysis}
\label{apd:statistics}

To ensure the reliability of our reported improvements, we conducted stability analysis by repeating experiments across five random seeds ($N=5$) for representative datasets from each domain. Table~\ref{tab:significance} reports the mean accuracy and standard deviation comparing Denser against the strongest baseline. We performed a one-tailed paired t-test to assess statistical significance. The results demonstrate that Denser maintains low variance across runs and that the performance improvements are statistically significant ($p < 0.05$) across all domains, confirming that the observed gains stem from the efficacy of our dual-density framework rather than stochastic variance in generation.

\begin{table}[!t]
\centering
\small
\resizebox{\columnwidth}{!}{
\begin{tabular}{lcccc}
\toprule
\textbf{Dataset} & \textbf{Metric} & \textbf{Self-Verification} & \textbf{Denser (Ours)} & \textbf{\textit{p}-value} \\
\midrule
GSM8K & Acc (\%) & 87.0 $\pm$ 0.42 & \textbf{88.2 $\pm$ 0.35} & 0.002$^{**}$ \\
LogiQA & Acc (\%) & 72.9 $\pm$ 0.58 & \textbf{74.1 $\pm$ 0.44} & 0.015$^{*}$ \\
MBPP & Pass@1 & 66.1 $\pm$ 0.65 & \textbf{67.3 $\pm$ 0.51} & 0.011$^{*}$ \\
MMLU & Acc (\%) & 80.4 $\pm$ 0.31 & \textbf{81.5 $\pm$ 0.28} & 0.004$^{**}$ \\
\bottomrule
\end{tabular}}
\caption{Statistical significance testing over 5 runs with different seeds. We compare Mean $\pm$ Standard Deviation against the strongest baseline. * indicates $p < 0.05$, ** indicates $p < 0.01$.}
\label{tab:significance}
\end{table}

\begin{figure}[!t]
    \centering
    \includegraphics[width=.93\linewidth]{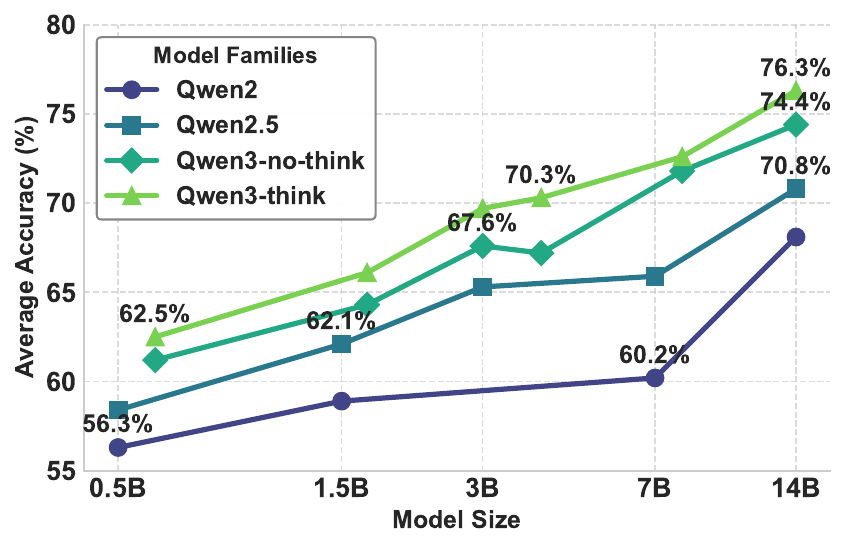}
    \caption{Performance across model sizes.}
    \label{fig:model_size}
\end{figure}

\subsection{Denser outperforms for all sizes of model.}
\label{apd:various_models}

To investigate the generalizability of Denser across parameter scales, we evaluate its performance on models of varying sizes: Qwen2 from 0.5B to 57B-A14B, Qwen2.5 from 0.5B to 14B, Qwen3 from 0.6B to 14B with both think and no-think mode. As illustrated in Figure~\ref{fig:model_size}, the results demonstrate that Denser consistently improves performance across all model sizes and architectures, with accuracy generally increasing with parameter count despite some non-linear scaling behaviors. Notably, the think variant of Qwen3 maintains a performance advantage over its no-think counterpart across all scales, with their gap varying at different parameter counts which highlights that Denser effectively leverages sophisticated reasoning capabilities regardless of model scale.

\section{More Case Studies}
\label{apd:case study}

Table~\ref{tab:extended_case_studies 2} to~\ref{tab:extended_case_studies 10} show the extended case study examples.

\begin{table*}[!t]
\centering
\small
\adjustbox{max width=\linewidth}{
\begin{tabular}{p{\linewidth}}
\toprule
\textbf{Case Study 2: Logical Reasoning} \\
\midrule
\textbf{Problem:} All lawyers are liars. Some liars are rich. Therefore, some lawyers are rich. Is this a valid argument? \\

\textbf{CoT:} Let me analyze this logical argument step by step. The first premise is "All lawyers are liars." This can be expressed as: If x is a lawyer, then x is a liar. The second premise is "Some liars are rich." This means that there exists at least one person who is both a liar and rich. Now, let's examine the conclusion: "Some lawyers are rich." This states that there exists at least one person who is both a lawyer and rich. To determine if the argument is valid, I need to check if the conclusion necessarily follows from the premises. From the first premise, we know that all lawyers are liars, so the set of lawyers is a subset of the set of liars. From the second premise, we know that the intersection of the set of liars and the set of rich people is non-empty. However, this doesn't guarantee that the intersection of lawyers and rich people is non-empty. The liars who are rich might not be lawyers. Consider a Venn diagram: we have lawyers (all of whom are liars), liars (some of whom are rich), and rich people. It's possible that the rich liars are not from the lawyer subset of liars. Therefore, the conclusion does not necessarily follow from the premises, and the argument is not valid. \\

\textbf{Denser (HD reasoning):}\\
Define sets:\\
L = lawyers\\
R = liars\\
S = rich people\\
P1: L $\subseteq$ R (All lawyers are liars)\\
P2: R $\cap$ S $\neq$ $\theta$ (Some liars are rich)\\
C: L $\cap$ S $\neq$ $\theta$ (Some lawyers are rich)

Check validity:\\
P1 $+$ P2 $\to$ C?\\
L $\subseteq$ R $\land$ R $\cap$ S $\neq \theta$ $\to$ L $\cap$ S $\neq \theta$?\\
Counterexample:\\
L = {a, b}\\
R = {a, b, c, d}\\
S = {c, e}\\
Here:\\
- L $\subseteq$ R (satisfied)\\
- R $\cap$ S = $\neq \theta$ (satisfied)\\
- But L $\cap$ S = $\theta$ (conclusion false)

Hence Invalid argument \\

\textbf{Denser (Final answer):} This argument is not valid. Let me analyze it using set theory:

Given:
- All lawyers are liars (All members of set L are in set R)
- Some liars are rich (The intersection of sets R and S is non-empty)

Conclusion:
- Some lawyers are rich (The intersection of sets L and S is non-empty)

While we know that lawyers are a subset of liars, and some liars are rich, this doesn't guarantee that any lawyers are among those rich liars.

Consider this counterexample:
- Lawyers = {Alice, Bob}
- Liars = {Alice, Bob, Charlie, David}
- Rich people = {Charlie, Eve}

In this example:
- All lawyers are liars (Alice and Bob are both liars)
- Some liars are rich (Charlie is both a liar and rich)
- But no lawyers are rich (Alice and Bob are not rich)

Since we can construct a counterexample where the premises are true but the conclusion is false, this argument is invalid. The conclusion doesn't necessarily follow from the premises.\\
\bottomrule
\end{tabular}}
\caption{Extended case study example 2}
\label{tab:extended_case_studies 2}
\end{table*}

\begin{table*}[!t]
\centering
\small
\adjustbox{max width=\linewidth}{
\begin{tabular}{p{\linewidth}}
\toprule
\textbf{Case Study 3: Mathematical Reasoning} \\
\midrule
\textbf{Problem:} Find the number of positive integers less than 1000 that are relatively prime to 210. \\

\textbf{CoT:} To solve this problem, I need to find the number of positive integers less than 1000 that are relatively prime to 210. First, let me find the prime factorization of 210. I'll start by dividing by small primes: 210 ÷ 2 = 105, so 2 is a factor. Then 105 ÷ 3 = 35, so 3 is a factor. Then 35 ÷ 5 = 7, so 5 is a factor. And 7 is prime. Therefore, 210 = 2 × 3 × 5 × 7. Now I need to use Euler's totient function. For any integer n with prime factorization $n = p_1^{a_1} \times p_2^{a_2} \times \ldots \times p_k^{a_k}$, we have $\phi(n) = n \times (1 - \frac{1}{p_1}) \times (1 - \frac{1}{p_2}) \times \ldots \times (1 - \frac{1}{p_k})$. So $\phi(210) = 210 \times (1 - \frac{1}{2}) \times (1 - \frac{1}{3}) \times (1 - \frac{1}{5}) \times (1 - \frac{1}{7}) = 210 \times \frac{1}{2} \times \frac{2}{3} \times \frac{4}{5} \times \frac{6}{7} = 210 \times \frac{1 \times 2 \times 4 \times 6}{2 \times 3 \times 5 \times 7} = 210 \times \frac{48}{210} = 48$. This gives us the count for one complete period of 210. Since we want numbers less than 1000, I need to consider how many complete periods of 210 fit into 1000, and handle the remainder separately. We have 1000 ÷ 210 = 4 remainder 160. So there are 4 complete periods, each contributing 48 numbers. For the remainder (numbers from 841 to 999), I need to count those relatively prime to 210. \\

\textbf{Denser (HD reasoning):} 
$210 = 2 \cdot 3 \cdot 5 \cdot 7$
$\phi(210) = 210 \cdot (1-\frac{1}{2})(1-\frac{1}{3})(1-\frac{1}{5})(1-\frac{1}{7}) = 210 \cdot \frac{1 \cdot 2 \cdot 4 \cdot 6}{2 \cdot 3 \cdot 5 \cdot 7} = 48$

$1000 = 4 \cdot 210 + 160$
Complete periods: $4 \cdot 48 = 192$

For remainder $[841, 999]$ (length 159):
Need $\gcd(n, 210) = 1$ for $n \in [841, 999]$
$\equiv n \in [1, 159] \pmod{210}$

Count using inclusion-exclusion:
$|A| = 159$
$|A_2| = \lfloor\frac{159}{2}\rfloor = 79$
$|A_3| = \lfloor\frac{159}{3}\rfloor = 53$
$|A_5| = \lfloor\frac{159}{5}\rfloor = 31$
$|A_7| = \lfloor\frac{159}{7}\rfloor = 22$

$|A_{2,3}| = \lfloor\frac{159}{6}\rfloor = 26$, $|A_{2,5}| = 15$, $|A_{2,7}| = 11$
$|A_{3,5}| = 10$, $|A_{3,7}| = 7$, $|A_{5,7}| = 4$

$|A_{2,3,5}| = 5$, $|A_{2,3,7}| = 3$, $|A_{2,5,7}| = 2$, $|A_{3,5,7}| = 1$
$|A_{2,3,5,7}| = 0$

Remainder count: $159 - (79+53+31+22) + (26+15+11+10+7+4) - (5+3+2+1) + 0 = 36$

Total: $192 + 36 = 228$ \\

\textbf{Denser (Final answer):} To find positive integers less than 1000 that are relatively prime to 210, I'll use Euler's totient function and inclusion-exclusion principle.

First, I find the prime factorization: $210 = 2 \times 3 \times 5 \times 7$.

Using Euler's totient function:
$\phi(210) = 210 \times (1-\frac{1}{2}) \times (1-\frac{1}{3}) \times (1-\frac{1}{5}) \times (1-\frac{1}{7}) = 210 \times \frac{48}{210} = 48$

Since $1000 = 4 \times 210 + 160$, there are 4 complete periods of 210, each contributing 48 relatively prime numbers, giving us $4 \times 48 = 192$.

For the remaining numbers from 841 to 999 (159 numbers), I use inclusion-exclusion to count those relatively prime to 210. This involves counting numbers divisible by 2, 3, 5, or 7, then applying the inclusion-exclusion principle.

After calculation, 36 numbers in the range [841, 999] are relatively prime to 210.

Therefore, the total count is $192 + 36 = 228$. \\

\bottomrule
\end{tabular}
}
\caption{Extended case study example 3}
\label{tab:extended_case_studies 3}

\end{table*}
\begin{table*}[!t]
\centering
\small
\adjustbox{max width=\linewidth}{
\begin{tabular}{p{\linewidth}}
\toprule
\textbf{Case Study 4: Mathematical Reasoning} \\
\midrule
\textbf{Problem:} In how many ways can 8 people be seated around a circular table such that 3 specific people must sit together? \\

\textbf{CoT:} I need to find the number of ways to seat 8 people around a circular table where 3 specific people must sit together. Let me call these 3 specific people A, B, and C. Since they must sit together, I can treat them as a single unit. So instead of arranging 8 individual people, I'm arranging 6 units around a circular table (the ABC group plus 5 other individuals). For circular arrangements, we fix one position to account for rotational symmetry. The number of ways to arrange n distinct objects in a circle is (n-1)!. So I have 6 units to arrange in a circle, which can be done in (6-1)! = 5! = 120 ways. But within the ABC group, the 3 people can be arranged among themselves. A, B, and C can be arranged in 3! = 6 different ways. Therefore, the total number of arrangements is 5! × 3! = 120 × 6 = 720. \\

\textbf{Denser (HD reasoning):} 
Treat 3 specific people as single unit.
Units: {ABC-group, P$_4$, P$_5$, P$_6$, P$_7$, P$_8$} = 6 units
Circular arrangement: $(6-1)! = 5! = 120$
Internal arrangement of ABC-group: $3! = 6$
Total: $5! \times 3! = 120 \times 6 = 720$ \\

\textbf{Denser (Final answer):} To solve this problem, I'll treat the 3 specific people who must sit together as a single unit.

Since the 3 specific people must sit together, I can consider them as one "super-person." This reduces the problem to arranging 6 units around a circular table: the group of 3 people plus the 5 other individuals.

For circular arrangements, we account for rotational symmetry by fixing one position. The number of ways to arrange 6 units in a circle is $(6-1)! = 5! = 120$.

However, within their group, the 3 specific people can be arranged among themselves in $3! = 6$ different ways.

Therefore, the total number of arrangements is $5! \times 3! = 120 \times 6 = 720$. \\

\bottomrule
\end{tabular}}
\caption{Extended case study example 4}
\label{tab:extended_case_studies 4}
\end{table*}

\begin{table*}[!t]
\centering
\small
\adjustbox{max width=\linewidth}{
\begin{tabular}{p{\linewidth}}
\toprule
\textbf{Case Study 5: Coding} \\
\midrule
\textbf{Problem:} Write an algorithm to find the longest increasing subsequence in an array of integers. Analyze its time complexity. \\

\textbf{CoT:} To find the longest increasing subsequence (LIS), I need to design an efficient algorithm. There are several approaches, but I'll use dynamic programming with binary search for optimal time complexity. The idea is to maintain an array that stores the smallest ending element of all increasing subsequences of length i+1 at index i. Let me call this array 'tails'. I'll also use binary search to efficiently find the position to update. Here's how the algorithm works: Initialize an empty array 'tails'. For each element in the input array, if the element is larger than all elements in 'tails', append it to 'tails'. Otherwise, find the smallest element in 'tails' that is greater than or equal to the current element and replace it. The length of 'tails' at the end gives us the length of the LIS. For the binary search, I need to find the leftmost position where I can place the current element. The time complexity analysis: For each of the n elements, I perform a binary search on the 'tails' array, which has at most n elements. Binary search takes O(log n) time. Therefore, the overall time complexity is O(n log n). The space complexity is O(n) for the 'tails' array. \\

\textbf{Denser (HD reasoning):} 
\begin{verbatim}
Algorithm: LIS_BinarySearch(arr)
tails = []
for x in arr:
    pos = binary_search_left(tails, x)
    if pos == len(tails):
        tails.append(x)
    else:
        tails[pos] = x
return len(tails)

binary_search_left(arr, target):
    left, right = 0, len(arr)
    while left < right:
        mid = (left + right) // 2
        if arr[mid] < target:
            left = mid + 1
        else:
            right = mid
    return left
\end{verbatim}
Invariant: `tails[i]` = min ending element of all inc. subseq. of length `i+1`
Time: O(n $\log$ n) - n elements $\times$ O($\log$ n) binary search
Space: O(n) \\

\textbf{Denser (Final answer):} Here's an efficient algorithm to find the longest increasing subsequence (LIS) using dynamic programming with binary search:

**Algorithm:**
\begin{verbatim}
def longest_increasing_subsequence(arr):
    tails = []
    
    for num in arr:
        # Binary search for the position to insert/replace
        left, right = 0, len(tails)
        while left < right:
            mid = (left + right) // 2
            if tails[mid] < num:
                left = mid + 1
            else:
                right = mid
        
        # If num is larger than all elements in tails, append it
        if left == len(tails):
            tails.append(num)
        else:
            # Replace the element at position 'left'
            tails[left] = num
    
    return len(tails)
\end{verbatim}

**Key Insight:** The `tails` array maintains the invariant that `tails[i]` stores the smallest ending element of all increasing subsequences of length `i+1`.

**Time Complexity:** O(n $\log$ n)
- We process each of the n elements once
- For each element, we perform binary search on the tails array (at most n elements)
- Binary search takes O($\log$ n) time

**Space Complexity:** O(n) for the tails array.

This algorithm is optimal for the LIS problem and significantly more efficient than the naive O(n$^2$) dynamic programming approach. \\

\bottomrule
\end{tabular}}
\caption{Extended case study example 5}
\label{tab:extended_case_studies 5}
\end{table*}

\begin{table*}[!t]
\centering
\small
\adjustbox{max width=\linewidth}{
\begin{tabular}{p{\linewidth}}
\toprule
\textbf{Case Study 6: Coding} \\
\midrule
\textbf{Problem:} Design a data structure that supports insert, delete, and getRandom operations in O(1) average time. \\

\textbf{CoT:} I need to design a data structure that supports three operations in O(1) average time: insert, delete, and getRandom. The key challenge is the getRandom operation, which requires accessing a random element efficiently. Let me think about this step by step. For insert and delete to be O(1), I need a hash-based structure. For getRandom to be O(1), I need to be able to access elements by index. The solution is to use a combination of an array (or list) and a hash map. The array will store the actual values, and the hash map will store the mapping from values to their indices in the array. For insert operation: Add the value to the end of the array and update the hash map with the value and its index. For delete operation: This is the tricky part. To maintain O(1) time, I can't just remove an element from the middle of the array (that would be O(n)). Instead, I'll swap the element to be deleted with the last element, then remove the last element. I also need to update the hash map accordingly. For getRandom operation: Generate a random index between 0 and the current size of the array, then return the element at that index. Let me write the code implementation with proper handling of edge cases and the hash map updates. \\

\textbf{Denser (HD reasoning):} 
\begin{verbatim}
Data Structure: RandomizedSet
- values: dynamic array
- indices: hashmap (value -> index)

insert(val):
    if val in indices: return false
    indices[val] = len(values)
    values.append(val)
    return true

delete(val):
    if val not in indices: return false
    idx = indices[val]
    last_val = values[-1]
    values[idx] = last_val
    indices[last_val] = idx
    values.pop()
    del indices[val]
    return true

getRandom():
    return values[random.randint(0, len(values)-1)]
\end{verbatim}
Time: O(1) avg for all ops
Space: O(n) \\

\textbf{Denser (Final answer):} I'll design a randomized set using a combination of a dynamic array and hash map to achieve O(1) average time for all operations.

**Data Structure Design:**
\begin{verbatim}
import random

class RandomizedSet:
    def __init__(self):
        self.values = []        # Dynamic array to store values
        self.indices = {}       # Hash map: value -> index in array
    def insert(self, val):
        if val in self.indices:
            return False
        # Add to end of array and update hash map
        self.indices[val] = len(self.values)
        self.values.append(val)
        return True
    def delete(self, val):
        if val not in self.indices:
            return False
        # Get index of element to delete
        idx = self.indices[val]
        last_element = self.values[-1]
        # Move last element to position of deleted element
        self.values[idx] = last_element
        self.indices[last_element] = idx
        # Remove last element and update hash map
        self.values.pop()
        del self.indices[val]
        return True
    def getRandom(self):
        return self.values[random.randint(0, len(self.values) - 1)]
\end{verbatim}
\\
\bottomrule
\end{tabular}
}
\caption{Extended case study example 6}
\label{tab:extended_case_studies 6}
\end{table*}

\begin{table*}[!t]
\centering
\small
\adjustbox{max width=\linewidth}{
\begin{tabular}{p{\linewidth}}
\toprule
\textbf{Case Study 7: Logical Reasoning} \\
\midrule
\textbf{Problem:} Determine if the following argument is valid using formal logic: "If it rains, then the ground gets wet. If the ground gets wet, then the plants grow. It is raining. Therefore, the plants are growing." \\

\textbf{CoT:} I need to analyze this argument using formal logic to determine its validity. Let me first identify the premises and conclusion, then translate them into logical statements. Let me define my propositional variables: Let R = "it rains", W = "the ground gets wet", and G = "the plants grow". Now I can translate the argument: Premise 1: "If it rains, then the ground gets wet" translates to R $\to$ W. Premise 2: "If the ground gets wet, then the plants grow" translates to W $\to$ G. Premise 3: "It is raining" translates to R. Conclusion: "Therefore, the plants are growing" translates to G. So the argument structure is: (R $\to$ W) $\land$ (W $\to$ G) $\land$ R $\vdash$ G. To check validity, I need to verify if the conclusion logically follows from the premises. I can use the transitivity of implication and modus ponens. From premises 1 and 2, using transitivity: R $\to$ W and W $\to$ G gives us R $\to$ G. From this derived statement R $\to$ G and premise 3 R, using modus ponens, we get G. Since I can derive the conclusion from the premises using valid inference rules, this argument is valid. \\

\textbf{Denser (HD reasoning):} 
Premises:
P1: R $\to$ W
P2: W $\to$ G  
P3: R
Conclusion: G

Proof:
1. R $\to$ W          (P1)
2. W $\to$ G          (P2)  
3. R $\to$ G          (1,2, transitivity)
4. R              (P3)
5. G              (3,4, modus ponens)

Hence Valid argument \\

\textbf{Denser (Final answer):} I'll analyze this argument using formal propositional logic.

**Translation to logical form:**
Let:
- R = "it rains"
- W = "the ground gets wet"  
- G = "the plants grow"

The argument becomes:
- Premise 1: R $\to$ W (If it rains, then the ground gets wet)
- Premise 2: W $\to$ G (If the ground gets wet, then the plants grow)
- Premise 3: R (It is raining)
- Conclusion: G (The plants are growing)

**Validity Check:**
I can prove the conclusion follows from the premises:

1. From premises 1 and 2, using transitivity of implication: If R $\to$ W and W $\to$ G, then R $\to$ G
2. From the derived statement R $\to$ G and premise 3 (R), using modus ponens: G

Since I can derive the conclusion G from the premises using valid logical inference rules (transitivity and modus ponens), this argument is **valid**.

The logical structure demonstrates a clear chain of reasoning where the conclusion necessarily follows from the premises. \\

\bottomrule
\end{tabular}}
\caption{Extended case study example 7}
\label{tab:extended_case_studies 7}
\end{table*}

\begin{table*}[!t]
\centering
\small
\adjustbox{max width=\linewidth}{
\begin{tabular}{p{\linewidth}}
\toprule
\textbf{Case Study 8: Logical Reasoning} \\
\midrule
\textbf{Problem:} Five friends (Alice, Bob, Carol, Dave, Eve) live on different floors (1-5) of an apartment building. Using these clues, determine who lives on which floor: (1) Alice lives two floors above Carol. (2) Bob lives on an even-numbered floor. (3) Dave lives on floor 3. (4) Eve lives on a higher floor than Alice. (5) No one lives on the same floor. \\

\textbf{CoT:} I need to solve this logic puzzle step by step using the given constraints. Let me work through each clue systematically. From clue 3, I know Dave lives on floor 3. This is a fixed point. From clue 1, Alice lives two floors above Carol. This means the possible combinations are: Carol on floor 1 and Alice on floor 3, Carol on floor 2 and Alice on floor 4, or Carol on floor 3 and Alice on floor 5. But Dave is already on floor 3, so Carol can't be on floor 3, and Alice can't be on floor 3. This eliminates the first option. So either Carol is on floor 2 and Alice on floor 4, or Carol is on floor 3 and Alice on floor 5. Wait, Carol can't be on floor 3 because Dave is there. So Carol must be on floor 2 and Alice on floor 4, or Carol could be on floor 1 and Alice on floor 3, but Alice can't be on floor 3 because Dave is there. Actually, let me reconsider: if Carol is on floor 1, Alice would be on floor 3, but Dave is on floor 3. If Carol is on floor 2, Alice would be on floor 4. If Carol is on floor 3, Alice would be on floor 5, but Dave is on floor 3. So Carol must be on floor 2, and Alice on floor 4. From clue 4, Eve lives on a higher floor than Alice. Since Alice is on floor 4, Eve must be on floor 5. From clue 2, Bob lives on an even-numbered floor. The even floors are 2 and 4. Alice is on floor 4, Carol is on floor 2, so Bob must be on floor 2. But Carol is on floor 2. This is a contradiction. Let me reconsider... \\

\textbf{Denser (HD reasoning):} 
Given: Floors 1,2,3,4,5 for Alice,Bob,Carol,Dave,Eve

C3: Dave $\to$ 3
C1: Alice = Carol + 2
C4: Eve > Alice  
C2: Bob $\in$ {2,4}

From C3: Dave = 3
From C1: (Carol,Alice) $\in$ {(1,3),(2,4),(3,5)}
Since Dave = 3: (Carol,Alice) $\in$ {(2,4),(3,5)}
But Dave = 3 $\neq$ Carol, so Carol $\neq$ 3
Therefore: (Carol,Alice) = (2,4)

From C4: Eve > 4, so Eve = 5
From C2: Bob $\in$ {2,4}, but Alice = 4, Carol = 2
Contradiction: Bob cannot occupy floor 2 or 4

Check (Carol,Alice) = (1,3):
Dave = 3, Alice = 3 $\to$ Contradiction

Recheck C1: Alice = Carol + 2
Only valid: Carol = 1, Alice = 3
But Dave = 3 $\to$ Impossible

No solution exists. \\

\textbf{Denser (Final answer):} I'll solve this step-by-step using logical deduction.

**Given constraints:**
1. Alice lives two floors above Carol
2. Bob lives on an even-numbered floor (2 or 4)
3. Dave lives on floor 3
4. Eve lives on a higher floor than Alice
5. Each person lives on a different floor

**Solution process:**

From constraint 3: Dave = floor 3

From constraint 1: Alice = Carol + 2
Possible combinations: (Carol=1, Alice=3), (Carol=2, Alice=4), or (Carol=3, Alice=5)

Since Dave occupies floor 3, Alice cannot be on floor 3, eliminating (Carol=1, Alice=3).
Since Dave occupies floor 3, Carol cannot be on floor 3, eliminating (Carol=3, Alice=5).
Therefore: Carol = floor 2, Alice = floor 4

From constraint 4: Eve > Alice = 4, so Eve = floor 5

From constraint 2: Bob must be on floor 2 or 4
But Carol = floor 2 and Alice = floor 4, so both even floors are occupied.

**Conclusion:** This puzzle has **no solution**. The constraints create a logical contradiction - Bob cannot be placed on any even-numbered floor because floors 2 and 4 are already occupied by Carol and Alice respectively. \\

\bottomrule
\end{tabular}}
\caption{Extended case study example 8}
\label{tab:extended_case_studies 8}
\end{table*}

\begin{table*}[!t]
\centering
\small
\adjustbox{max width=\linewidth}{
\begin{tabular}{p{\linewidth}}
\toprule
\textbf{Case Study 9: General QA (MMLU} \\
\midrule
\textbf{Problem:} A 2 kg block slides down a frictionless inclined plane that makes a 30$\degree$ angle with the horizontal. What is the acceleration of the block down the plane? (A) 4.9 m/s$^2$ (B) 9.8 m/s$^2$ (C) 8.5 m/s$^2$ (D) 19.6 m/s$^2$ \\

\textbf{CoT:} I need to find the acceleration of a block sliding down a frictionless inclined plane. Let me set up the problem systematically. The forces acting on the block are: the gravitational force (weight) mg acting vertically downward, and the normal force from the inclined plane acting perpendicular to the surface. Since the plane is frictionless, there's no friction force. I need to resolve the gravitational force into components parallel and perpendicular to the inclined plane. The component of gravitational force parallel to the plane (down the slope) is mg sin $\theta$, where $\theta$ is the angle of inclination. The component perpendicular to the plane is mg cos $\theta$, which is balanced by the normal force. Applying Newton's second law along the plane: The net force down the plane is mg sin $\theta$. Using F = ma, we get: ma = mg sin $\theta$, so a = g $\sin$ $\theta$. Given that g = 9.8 m/s$^2$ and $\theta$ = 30$\degree$, I calculate: a = 9.8 × $\sin$(30$\degree$) = 9.8 × 0.5 = 4.9 m/s$^2$. Notice that the mass of the block (2 kg) doesn't affect the acceleration - this is a characteristic feature of motion under gravity alone. \\

\textbf{Denser (HD reasoning):} 
Forces on block:
- Weight: mg (vertical)
- Normal: N ($\bot$ to plane)
- Friction: 0 (frictionless)

Force components:
$\parallel$ to plane: mg sin $\theta$ (down)
$\bot$ to plane: mg cos $\theta$ (balanced by N)

Newton's 2nd law:
ma = mg sin $\theta$
a = g sin $\theta$ = 9.8 sin(30$\degree$) = 9.8 × 0.5 = 4.9 m/s$^2$

Answer: (A) \\

\textbf{Denser (Final answer):} To find the acceleration of the block down the inclined plane, I need to analyze the forces and apply Newton's second law.

**Force Analysis:**
The weight (mg) acts vertically downward. I'll resolve this into components:
- Parallel to plane: mg sin $\theta$ (down the slope)
- Perpendicular to plane: mg cos $\theta$ (balanced by normal force)

Since the plane is frictionless, the only unbalanced force is the component parallel to the plane.

**Applying Newton's Second Law:**
Along the inclined plane: F$_{net}$ = ma
mg sin $\theta$ = ma
Therefore: a = g sin $\theta$

**Calculation:**
Given: $\theta$ = 30$\degree$, g = 9.8 m/s$^2$
a = 9.8 × sin(30$\degree$) = 9.8 × 0.5 = 4.9 m/s$^2$

Note that the mass (2 kg) cancels out, so the acceleration is independent of the block's mass.

**Answer: (A) 4.9 m/s$^2$** \\

\bottomrule
\end{tabular}}
\caption{Extended case study example 9}
\label{tab:extended_case_studies 9}
\end{table*}

\begin{table*}[!t]
\centering
\small
\adjustbox{max width=\linewidth}{
\begin{tabular}{p{\linewidth}}
\toprule
\textbf{Case Study 10: General QA} \\
\midrule
\textbf{Problem:} Could a professional basketball player dunk while standing on the shoulders of an average-height person? \\

\textbf{CoT:} To answer this question, I need to consider the heights involved and what's required for dunking. Let me break this down systematically. First, let me establish the relevant measurements: A standard basketball hoop is 10 feet (3.05 meters) above the ground. The average height of an adult male in the US is approximately 5'9" (1.75 meters). A professional basketball player is typically much taller - let's use an average NBA player height of about 6'7" (2.01 meters). For a successful dunk, a player needs their hand to reach significantly above the rim to actually slam the ball down. This typically requires the hand to be at least 10'6" to 11' above the ground (3.2 to 3.35 meters). Now, if an NBA player stands on the shoulders of an average person: The average person's shoulder height is roughly at their total height minus their head height, so about 5'4" (1.63 meters) from the ground. Adding the NBA player's height: 5'4" + 6'7" = 11'11" (3.63 meters). This would put the NBA player's head at nearly 12 feet above ground, meaning their hand could easily reach 13+ feet (4+ meters). This is far more than enough to dunk comfortably. So yes, a professional basketball player could definitely dunk while standing on someone's shoulders. \\

\textbf{Denser (HD reasoning):} 
Required for dunk: hand $\geq$ 10.5 ft above ground
Basketball hoop: 10 ft
Average person height: 5'9" (1.75m)
Shoulder height $\approx$ 5'4" (1.63m)
NBA player avg height: 6'7" (2.01m)

Standing configuration:
Shoulder height + NBA player height
= 5'4" + 6'7" = 11'11" (3.63m)
NBA player's hand reach $\approx$ 13+ ft (4m+)

13 ft $>\!\!>$ 10.5 ft $\to$ Yes, easily possible \\

\textbf{Denser (Final answer):} Yes, a professional basketball player could easily dunk while standing on the shoulders of an average-height person.

**Analysis:**
- Basketball rim height: 10 feet
- Dunking requires hand to reach $\sim$10.5 feet (to get above rim)
- Average adult shoulder height: $\sim$5'4"
- Average NBA player height: $\sim$6'7"

**Combined reach:**
Standing on shoulders: 5'4" + 6'7" = 11'11"
This puts the player's hands at approximately 13+ feet above ground.

Since 13+ feet far exceeds the 10.5 feet needed for dunking, this would not only be possible but quite easy. The player would have over 2 feet of clearance above the rim.

**Conclusion:** Not only is this possible, but the basketball player would have an enormous amount of extra height, making it a very comfortable dunk. \\

\bottomrule
\end{tabular}}
\caption{Extended case study example 10}
\label{tab:extended_case_studies 10}
\end{table*}

\end{document}